\documentclass{article}

\usepackage{amsmath,amsfonts,bm}

\def\eqref#1{equation~\ref{#1}}
\def\1{\bm{1}}

\DeclareMathAlphabet{\mathsfit}{\encodingdefault}{\sfdefault}{m}{sl}
\SetMathAlphabet{\mathsfit}{bold}{\encodingdefault}{\sfdefault}{bx}{n}

\usepackage[utf8]{inputenc} % allow utf-8 input
\usepackage[T1]{fontenc}    % use 8-bit T1 fonts
\usepackage{url}            % simple URL typesetting
\usepackage{booktabs}       % professional-quality tables
\usepackage{amsfonts}       % blackboard math symbols
\usepackage{nicefrac}       % compact symbols for 1/2, etc.
\usepackage{microtype}      % microtypography
\usepackage{amsmath}
\usepackage{graphicx}
\usepackage[colorlinks=true, allcolors=blue]{hyperref}
\usepackage[dvipsnames,table,xcdraw]{xcolor}
\usepackage{xspace}
\usepackage{multicol}
\usepackage{wrapfig}
\usepackage{multirow}
\usepackage{booktabs} % For prettier tables
\usepackage{makecell} % For line breaks in table cells
\usepackage{arydshln}
\usepackage{enumitem}

\usepackage{hyperref}

\usepackage[disable]{todonotes}

\usepackage[accepted]{icml2026}

\usepackage{amsmath}
\usepackage{amssymb}
\usepackage{mathtools}
\usepackage{amsthm}

\usepackage[capitalize,noabbrev]{cleveref}

\theoremstyle{plain}

\theoremstyle{definition}

\theoremstyle{remark}

\usepackage[disable]{todonotes}

\icmltitlerunning{Neural Concept Verifier: Scaling Prover-Verifier Games via Concept Encodings}

\newcommand{\methodfull}{Neural Concept Verifier\xspace}
\newcommand{\method}{NCV\xspace}

\definecolor{btcolor}{RGB}{220,20,60}

\newcommand{\eg}{\textit{e.g.}} 

\newcommand{\ie}{\textit{i.e.}} 

\newcommand{\cf}{\textit{cf.}~}

\newcommand{\std}[1]{\mbox{\scriptsize$\pm\,#1$}}

\newcounter{mycomment}

\begin{document}

\twocolumn[
  \icmltitle{Neural Concept Verifier: \\ Scaling Prover-Verifier Games via Concept Encodings}



  \icmlsetsymbol{equal}{*}

  \begin{icmlauthorlist}
    \icmlauthor{Berkant Turan}{zib}
    \icmlauthor{Suhrab Asadulla}{zib}
    \icmlauthor{David Steinmann}{tuda}
    \icmlauthor{Kristian Kersting}{tuda,hessian,lab1141,dfki}
    \icmlauthor{Wolfgang Stammer}{mpi}
    \icmlauthor{Sebastian Pokutta}{zib,tubmath}
  \end{icmlauthorlist}

  \icmlaffiliation{zib}{Zuse Institute Berlin, Germany}
  \icmlaffiliation{tuda}{AI \& ML Lab, Computer Science Department, TU Darmstadt}
  \icmlaffiliation{hessian}{Hessian Center for AI (hessian.AI)}
  \icmlaffiliation{lab1141}{Lab1141}
  \icmlaffiliation{dfki}{German Research Center for AI (DFKI)}
  \icmlaffiliation{tubmath}{Institute of Mathematics, Technische Universit\"at Berlin, Germany}
  \icmlaffiliation{mpi}{Max Planck Institute for Informatics, SIC}

  \icmlcorrespondingauthor{Berkant Turan}{turan@zib.de}

  \icmlkeywords{Interpretability, Prover-Verifier Games, Concept Bottleneck Models, Concept Explanation, XAI}

  \vskip 0.3in
]



\printAffiliationsAndNotice{}  % no special notice (required even if empty)

\begin{abstract}
% While \textit{Prover-Verifier Games} (PVGs) offer a promising path toward verifiability in nonlinear classification models, they have not yet been applied to complex inputs such as high-dimensional images. 
% Conversely, expressive \textit{concept representations} effectively allow to translate such data into interpretable concepts but are often utilised in the context of low-capacity linear predictors.
% In this work, we push towards real-world verifiability by combining the strengths of both approaches. We introduce \textit{Neural Concept Verifier (NCV)}, a unified framework combining PVGs for formal verifiability with concept encodings to handle complex, high-dimensional inputs in an interpretable way. NCV achieves this by utilizing recent minimally supervised concept discovery models to extract structured concept encodings from raw inputs. A \textit{prover} then selects a subset of these encodings, which a \textit{verifier}, implemented as a nonlinear predictor, uses exclusively for decision-making.
% Our evaluations show that NCV outperforms CBM and pixel-based PVG classifier baselines on high-dimensional, logically complex datasets and also helps mitigate shortcut behavior. Overall, we demonstrate NCV as a promising step toward concept-level, verifiable AI.

While \textit{Prover-Verifier Games} (PVGs) offer a promising path toward verifiability in nonlinear classification models, they have not yet been applied to complex inputs such as high-dimensional images. 
Conversely, expressive \textit{concept encodings} effectively allow to translate such data into interpretable concepts but are often utilised in the context of low-capacity linear predictors.
In this work, we push towards real-world verifiability by combining the strengths of both approaches. We introduce \textit{Neural Concept Verifier (NCV)}, a unified framework combining PVGs for formal verifiability with concept encodings to handle complex, high-dimensional inputs in an interpretable way. NCV achieves this by utilizing recent minimally supervised concept discovery models to extract structured concept encodings from raw inputs. A \textit{prover} then selects a subset of these encodings, which a \textit{verifier}, implemented as a nonlinear predictor, uses exclusively for decision-making.
Our evaluations show that NCV outperforms classic concept-based models and pixel-based PVG classifier baselines on high-dimensional, logically complex datasets and helps mitigate shortcut behavior. Overall, we demonstrate NCV as a promising step toward concept-level, verifiable AI.
\end{abstract}

\renewcommand{\figureautorefname}{Fig.\xspace}
\renewcommand{\tableautorefname}{Tab.\xspace}
\renewcommand{\sectionautorefname}{Sec.\xspace}
\renewcommand{\equationautorefname}{Eq.\xspace}
\renewcommand{\appendixautorefname}{Suppl.\xspace}
\renewcommand{\subsectionautorefname}{Sec.\xspace}
\renewcommand{\subsubsectionautorefname}{Sec.\xspace}
\newcommand{\algorithmautorefname}{Alg.\xspace}

\section{Introduction}

\begin{figure}
    \centering
    \includegraphics[width=0.9\linewidth]{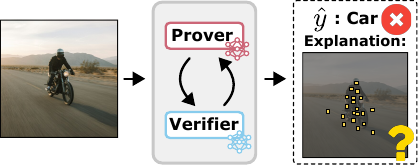}
    \caption{Challenges of Prover-Verifier Games (PVGs) in image classification: (i) It is non-trivial to scale up for high-dimensional data. (ii) Furthermore, the learned explanation masks on the pixel level remain difficult for humans to understand.}
    \label{fig:motivation}    
\end{figure}

Deep learning has achieved remarkable predictive performances, but often at the expense of \textit{interpretability and trustworthiness} \citep{rudin2019stop}. However, particularly in high-stakes applications, it is critical that models provide \textit{verifiable justifications} for their decisions \citep{irving2018aisafetydebate,fok23}. \textit{Prover-Verifier Games} (PVGs), introduced by \citet{anil2021learning}, formalize such justifications via a game-theoretic approach: a prover provides evidence to convince a verifier, who accepts only verifiable proofs. A prominent instantiation of PVGs is the \textit{Merlin-Arthur Classifier}~\citep{pmlr-v238-waldchen24a}, a classifier guided by cooperative and adversarial provers, offering formal interpretability guarantees through information-theoretic bounds. Specifically, identified features are provably relevant for a given output, rather than mere correlations as in typical feature explanation methods \citep{adebayo2018sanity, zhou2022feature}. 
However, the Merlin-Arthur framework faces significant scalability challenges when applied to high-dimensional real-world data, as explanations based on raw pixels are both computationally difficult to optimize and offer limited human understandability~(\cf \autoref{fig:motivation}; \citet{pmlr-v238-waldchen24a}).

Concurrently, \textit{Concept Bottleneck Models} (CBMs) have emerged as a powerful framework for interpretable machine learning, structuring predictions through intermediate interpretable concept encodings \citep{koh2020concept,StammerSK21}. Despite their advantages, CBMs typically employ \textit{linear classifiers} on top of concept encoding layers, thereby potentially restricting expressivity and failing on tasks requiring nonlinear interactions among concepts (e.g., XOR problems, counting or permutation invariance, \cf \autoref{app:linear}) \citep{mahinpei2021promises, kimura2024permutation, pmlr-v97-lee19d}.

In this work, we combine the best of both worlds by introducing the \textbf{Neural Concept Verifier (NCV)}, a novel framework integrating concept-based representations into PVGs in the form of Merlin-Arthur Classifiers. NCV shifts the prover–verifier interaction from the image level to a structured, symbolic concept level, overcoming both the scalability limitations encountered by Merlin-Arthur Classifiers in high-dimensional settings and the expressivity constraints inherent to linear CBMs. We demonstrate through extensive evaluations that NCV successfully scales PVGs to complex, high-dimensional classification tasks. At the same time, NCV enables verifiable, performant nonlinear classifiers on top of concept extractors, effectively narrowing the interpretability–accuracy gap in linear CBMs. Lastly, our framework improves robustness to shortcut learning, thereby enhancing the generalizability and trustworthiness of predictions, particularly in high-stakes applications.

Our summarized contributions are: (i) We propose \textit{Neural Concept Verifier} (\method), a framework combining concept-based models with Prover–Verifier Games (PVGs). (ii) \method scales PVGs to high-dimensional image data by operating on compact concept encodings. (iii) It enables expressive yet interpretable classification via sparse, nonlinear reasoning over concepts. (iv) We validate \method on synthetic and real-world benchmarks, demonstrating strong accuracy and verifiability. (v) We highlight that \method improves generalization under spurious correlations. 
The remainder of the paper is structured as follows. We begin with a review of related work, followed by an introduction and formal description of \method. We then provide a comprehensive evaluation and, finally, discuss our findings and conclude the paper.

\section{Background}

\textbf{Prover-Verifier Games.}
PVGs were introduced by \citet{anil2021learning} as a game-theoretic framework to encourage learning agents to produce \textit{testable} justifications through interactions between an untrusted prover and a trusted verifier. Their work showed that, under suitable conditions, the verifier can learn robust decision rules even when the prover actively attempts to persuade it of arbitrary outputs. \citet{pmlr-v238-waldchen24a} extended this idea to the \textit{Merlin-Arthur Classifier} (MAC), which provides formal interpretability guarantees by bounding the mutual information between the selected features and the ground-truth label. Recently, \citet{Kirchner24PV} applied a PVG-inspired approach to improve legibility of Large Language Model (LLM) outputs 
and \citet{amit2024models} introduced \textit{self-proving models} that leverage interactive proofs to formally verify the correctness of model outputs. PVG-style setups have also been explored in safety-focused learning protocols~\citep{irving2018aisafetydebate, browncohen2024scalable, gluch2024goodbaduglywatermarks}. These developments reflect a broader trend of utilising multi-agent learning~\citep{PruthiBDSCLNC22,schneider2023reflective,Du23multiagent,Nair23,stammer2024learning}. Our work builds on this line of research by embedding PVGs into a concept-based classification framework to address the dimensionality bottleneck that limits their performance in high-dimensional settings. 

\textbf{Concept Representations for Interpretability.}
The introduction of Concept Bottleneck Models (CBMs) \citep{koh2020concept,espinosa2022concept} and concept-based explanations~\citep{kim2018interpretability,crabbe2022concept,poeta2023concept,lee2025neural} was an important moment in the growing interest in AI interpretability research. The particular appeal of CBMs lies in the promise of interpretable predictions and a controllable, structured interface for human interactions \citep{StammerSK21}. While initial CBMs relied on fully supervised concept annotations, subsequent research has relaxed this requirement by leveraging pretrained vision-language models like CLIP for concept extraction \citep{bhalla2024interpreting, yang2023language, oikarinen2023label, panousis2024coarse, steinmann2025object}, or employing fully unsupervised concept discovery methods \citep{ghorbani2019towards, stammer2024neural, schut2025bridging}. Post-hoc variants instead attach concept-based components on top of pretrained black-box predictors~\citep{yuksekgonul2023posthoc, pmlr-v280-wang25b}, while complementary methods statistically test the importance of semantic concepts for an existing classifier post hoc~\citep{NEURIPS2024_8c1df815}.

While most works have focused on the bottleneck itself by reducing supervision requirements, mitigating concept leakage, or dynamically expanding the concept space, several recent approaches also enrich the classifier component using nonlinear or symbolic predictors, including concept-based memory reasoning~\citep{debot2024interpretable}, neural-symbolic reasoning~\citep{barbiero2023interpretable}, and causal concept models~\citep{dominici2024causal,de2025causally}. A complementary information-theoretic perspective is taken by Variational Information Pursuit~\citep{chattopadhyay2023variational}, which sequentially selects queries to maximise mutual information rather than producing a single sparse selection over a fixed concept vocabulary (see \autoref{app:vip} for a detailed comparison).
NCV is complementary to these methods: rather than proposing a new classifier architecture, it wraps \textit{any} concept-level predictor in a prover--verifier game that enforces sparse, per-sample concept selection and evaluates predictions under competing subsets. In contrast to Sparse Autoencoders~\citep{cunningham2023sparse}, which decompose internal activations of black-box models into interpretable features post hoc, \method operates over pre-defined, grounded concept vocabularies and enforces sparse selection through adversarial training during learning.

\begin{figure*}[t!]
    \centering
    \includegraphics[width=.82\linewidth]{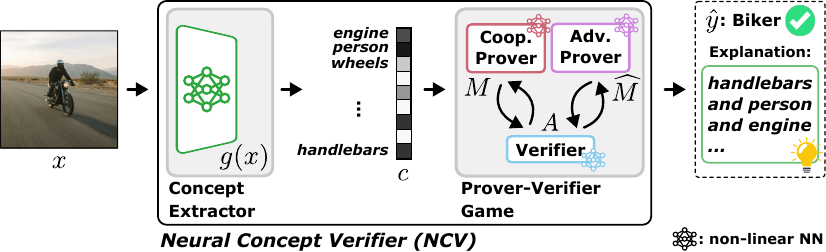}
    \caption{Overview of the Neural Concept Verifier (NCV). The input image is first processed by a concept extractor to produce symbolic concept encodings. A prover–verifier game is then played over these encodings: a cooperative prover selects a sparse concept subset supporting the true class, while an adversarial prover selects misleading concepts. Finally, the nonlinear verifier makes a prediction based only on these selected concepts, ensuring verifiable and robust classification.}
    \label{fig:method}
\end{figure*}

\textbf{Shortcut Learning.} Independent of their interpretability, training deep models can often lead to unwanted artifacts and side effects. Shortcut learning describes the problem of models learning to rely on unwanted and unintended features to resolve a task \citep{geirhos2020shortcut}. This is a common problem when training purely deep learning models~\citep{lapuschkin2019unmasking,SchramowskiSTBH20} or neuro-symbolic models~\citep{MarconatoTVP23,Bortolotti25nesy}, and, if not taken care of, can lead to predictions being right for the wrong reasons~\citep{ross2017right}. There have been various approaches to tackle this problem, from careful dataset curation \citep{ahmed2021discovery} to modified model training \citep{FriedrichSSK23b}, \textit{cf.} \citet{steinmann2024navigating} for a comprehensive overview. While \method is not specifically designed to mitigate shortcuts, we show that our setup can intrinsically mitigate their impact on the data.

\section{Neural Concept Verifier (NCV)}\label{sec:ncv}
In this section, we introduce the \textit{Neural Concept Verifier} (NCV), a framework that combines concept-based representations with the Merlin-Arthur prover-verifier paradigm~\citep{pmlr-v238-waldchen24a}. NCV trains nonlinear classifiers whose predictions provably rely on sparse subsets of high-level concepts, with information-theoretic guarantees formalized through completeness and soundness criteria (\cf \autoref{app:theory}). In contrast to the original Merlin-Arthur classifier (MAC), which operates on raw pixel features and struggles to scale to high-dimensional inputs, NCV performs the prover–verifier interaction directly in concept space. This shift enables optimization on complex datasets and grounds PVG explanations in human-interpretable concepts.

\method consists of two main components (\cf \autoref{fig:method}): (i) a weakly supervised \textit{concept extractor} that transforms input data into interpretable concept encodings, and (ii) a nonlinear MAC that selects sparse concepts for the final predictions. After providing background notations and a high-level overview of \method, we present detailed descriptions of each main component as well as training and inference details.

\subsection{Problem Setup and Notation}
\label{sec:setup}
Let \(\mathcal{X} \in \mathbb{R}^{N \times D}\) denote a dataset of \(N\) inputs (e.g., images), each of dimension \(D\), and let \(\mathcal{Y} \in \{1,\dots,K\}^N\) be the corresponding class labels for \(K\) classes. We assume that the pairs \((x,y)\) are drawn \textit{i.i.d.}\ from an unknown data distribution \(\mathcal{D}\) over \(\mathbb{R}^D \times \{1,\dots,K\}\), and that \((\mathcal{X},\mathcal{Y})\) correspond to a finite sample from \(\mathcal{D}\).
For the verifier, we consider an extended prediction space \(\{1,\dots,K,\bot\}\) with an additional \emph{rejection class} \(\bot\), allowing the classifier to abstain from a decision when uncertain, which is crucial for enforcing interpretability guarantees in adversarial setups~\citep{pmlr-v238-waldchen24a}.
Our overall goal is to learn a model \(f : \mathbb{R}^D \rightarrow \{1,\dots,K,\bot\}\) that maps an input \(x \in \mathbb{R}^D\) to a prediction \(\hat{y} \in \{1,\dots,K,\bot\}\) that is based on a small, interpretable subset of high-level concepts.

\subsection{The NCV Framework}
\method decomposes \(f\) into the following components (\cf \autoref{fig:method} for an illustrative overview):
\begin{enumerate}
    \item A \textbf{concept extractor} \(g: \mathcal{X} \rightarrow \mathcal{C}\), which maps each input \(x\) to a high-level concept encoding \(\mathbf{c} \in \mathbb{R}^C\), where \(C \in \mathbb{N}\) is the number of discovered concepts.
    \item A pair of \textbf{provers} \(M, \widehat{M}: \mathcal{C} \rightarrow \{0,1\}^C\) that produce sparse binary masks selecting \(m\) concepts each. \(M\) (Merlin, \textit{cooperative prover}) aims to help classification; \(\widehat{M}\) (Morgana, \textit{adversarial prover}) aims to mislead, which is crucial for overall robustness.
    \item A nonlinear \textbf{verifier} (Arthur) \(A: \mathbb{R}^C \rightarrow \mathcal{Y}\), which predicts a label based only on the masked concepts.
\end{enumerate}

The three modules are trained jointly, where their interaction encourages the verifier to rely only on robust, informative concept features.
Let us now provide details on these individual components.

\subsection{Concept Extraction}\label{sec:concept_extractors}
The concept extractor $g: \mathcal{X} \to \mathbb{R}^C$ transforms raw input data into interpretable, high-level concept representations, where each dimension corresponds to a semantically meaningful concept. The resulting concept encoding $\mathbf{c} \in \mathbb{R}^C$ serves as the input to the PVG. 

While conceptually simple, a careful combination of PVGs and concept-based models is necessary. The concept extractor must satisfy three key requirements: interpretability, expressiveness, and modularity. Concepts should correspond to human-understandable features that can serve as meaningful explanations, the concept space should capture sufficient information for the target task without creating information bottlenecks, and the extractor should operate independently of the prover-verifier components.
Unlike traditional concept extractor approaches that enforce sparsity constraints directly on $\mathbf{c}$, our framework delegates sparsity to the prover-verifier interaction. This allows the concept space to remain dense and expressive while achieving interpretable sparsity through downstream concept selection.
Further, NCV can accommodate different supervision paradigms for concept extraction:  supervised methods that leverage predefined concept vocabularies, self-supervised methods that exploit multi-modal correspondences (e.g., vision-language alignment), or unsupervised methods that discover latent conceptual structures can all be used as concept extractors.

Overall, NCV requires that $g$ produces consistent, interpretable encodings while maintaining sufficient information for accurate classification. In our evaluations, we instantiate NCV's concept extractor via the recent unsupervised, object-centric NCB framework~\citep{stammer2024neural} and the multi-modal, CLIP-based SpLiCE~\citep{bhalla2024interpreting}. Operating in concept space rather than raw input space provides: (i) scalability through dimensionality reduction and (ii) explanations based on human-interpretable concepts.

\subsection{Verifiable Classification via Merlin-Arthur Classifiers}\label{sec:merlin-arthur}

The second core component of \method is a interpretable classifier of the Merlin-Arthur setup \citep{pmlr-v238-waldchen24a} originally inspired by Interactive Proof Systems~\citep{goldwasserproofs}. This setup generally formalizes the idea of proving that a classification decision is supported by a sparse and informative set of features. 

In \method, specifically, two competing provers, \textbf{Merlin} and \textbf{Morgana}, select concept subsets either to support or mislead classification, respectively. The verifier, Arthur, then makes predictions based solely on these masked concepts without knowledge of the prover's intent. Formally, each prover outputs a sparse binary mask with $m$ active entries, producing Merlin's cooperative
subset $S = M(\mathbf{c}) \odot \mathbf{c}$ and Morgana's adversarial subset $\widehat S = \widehat{M}(\mathbf{c}) \odot \mathbf{c}$, where $\mathbf{c}$ is the concept encoding from extractor
$g$, and $\odot$ denotes element-wise masking. Notably, all three modules are differentiable nonlinear
models. 

This interactive setup enables two key metrics: \textit{completeness} and \textit{soundness}. With Merlin's subset $S$ and Morgana's subset $\widehat{S}$, we define
\begin{align}
    \text{Completeness} \;&=\; \mathbb{P}_{(x,y)\sim\mathcal{D}}\!\big[A(S)=y\big], \\
    \text{Soundness} \;&=\; \mathbb{P}_{(x,y)\sim\mathcal{D}}\!\big[A(\widehat S)\in\{y,\bot\}\big],
\end{align}
where $\bot$ denotes the rejection class. Intuitively, completeness measures how often Arthur can recover the true label from Merlin's sparse, helpful concepts (similar to standard accuracy), while soundness measures how often Arthur can avoid committing to a wrong label under Morgana's misleading subset, either by staying correct or abstaining (\cf \autoref{app:theory} for the theoretical interpretation).

Furthermore, sparsity has recently become central in concept-based models, as large concept spaces require sparse predictions for interpretability. Prior work \citep{bhalla2024interpreting, de2025towards} achieves this by regularizing the concept space itself, restricting the number of active concepts before training the classifier. In contrast, \method keeps the concept space fully expressive and enforces sparsity only in the concepts passed to the classifier.

\subsection{Information-Theoretic Interpretation}\label{sec:theory}

The completeness and soundness defined in \autoref{sec:merlin-arthur} are not just empirical metrics; they carry an information-theoretic interpretation grounded in the Merlin-Arthur classifier theory of \citet{pmlr-v238-waldchen24a}, instantiated class-wise in concept space. We summarize the key statements here and defer all formal derivations, full precision bounds, and a detailed discussion of limitations to \autoref{app:theory}.

\paragraph{Concept-space certificates.}
Since the prover $M$ from \autoref{sec:merlin-arthur} acts on the concept encoding, we slightly abuse notation and write $M(x) := M(g(x)) \in \{0,1\}^C$. Following the partial-input formalism of \citet{pmlr-v238-waldchen24a}, we write $M(x') \subseteq x$ for the event that $g(x)_i = g(x')_i$ at every index $i$ with $M(x')_i = 1$, i.e., $x$ and $x'$ agree on every concept that Merlin selects from $x'$. For a fixed class $k \in \{1,\dots,K\}$, define the one-vs-rest \emph{labeling function} $Y_k\!: \{1,\dots,K\} \to \{-1, +1\}$ by $Y_k(y) := +1$ if $y = k$ and $-1$ otherwise. We assume the label $y$ is determined by the input $x$. Imperfect precision then reflects insufficiency of the selected concepts rather than intrinsic label noise.

\paragraph{Class-wise precision and mutual information.}
For a fixed class $k$, the \emph{average precision} of $M$ is
\begin{equation*}
    \Pr_{\mathcal{D}}^{(k)}(M) := \mathbb{E}_{(x',y')}\!\big[\mathbb{P}_{(x,y)}\big[Y_k(y) = Y_k(y') \mid M(x') \subseteq x\big]\big].
\end{equation*}
Applying the binary mutual-information bound of \citet{pmlr-v238-waldchen24a} with $Y_k(y)$ as the binary label, we obtain a class-wise lower bound on the mutual information between the class-$k$ label and Merlin's certificate:
\begin{multline}
    \mathbb{E}_{(x',y')\sim\mathcal{D}}\!\big[I_{(x,y)\sim\mathcal{D}}\big(Y_k(y);\, M(x') \subseteq x\big)\big] \\
    \;\ge\; H_{y\sim\mathcal{D}}(Y_k(y)) - H_b\!\big(\Pr_{\mathcal{D}}^{(k)}(M)\big),
    \label{eq:mi-bound-multiclass}
\end{multline}
where $H_b\!:[0,1] \to [0,1]$ is the binary entropy and $M(x') \subseteq x$ in the mutual information denotes the binary containment variable for fixed $x'$.

\paragraph{Observable completeness and soundness.}
A single trained triple $(M, \widehat{M}, A)$ induces a pair of class-wise errors for every $k$:
\begin{align*}
    \varepsilon_c^{(k)} &:= \mathbb{P}_{(x,y)\sim\mathcal{D}}\!\big[A\big(M(g(x))\odot g(x)\big) \neq k \mid y = k\big], \\
    \varepsilon_s^{(k)} &:= \mathbb{P}_{(x,y)\sim\mathcal{D}}\!\big[A\big(\widehat{M}(g(x))\odot g(x)\big) = k \mid y \neq k\big],
\end{align*}
measuring how often Arthur fails to recognise class $k$ from Merlin's selection and how often Morgana flips Arthur into predicting $k$ on a non-$k$ input; these refine the aggregate completeness and soundness of \autoref{sec:merlin-arthur}, and \autoref{app:theory} gives the worst-class variant of \citet{pmlr-v238-waldchen24a} that enters the bound. Under three further conditions, namely a bounded asymmetric feature correlation $\kappa_k \geq 1$ and a relative success rate $\alpha_k > 0$ comparing Morgana's strength to Merlin's, together with the class imbalance $B_k \geq 1$ (directly estimable from label frequencies), $\Pr_{\mathcal{D}}^{(k)}(M)$ is lower-bounded in $\varepsilon_c^{(k)}$ and $\varepsilon_s^{(k)}$ (see \autoref{app:ncv-instantiation}). When both errors are small, Merlin's sparse concept selection almost determines the class.

\paragraph{Faithfulness and scope.}
We call an explanation \emph{faithful} (see \autoref{app:faithfulness}) when Arthur, with high probability, classifies correctly from Merlin's sparse selection and is not driven to a wrong prediction by Morgana; under the precision-bound conditions this yields the guarantee of \autoref{eq:mi-bound-multiclass}. The bound holds under two assumptions, $\kappa_k$ and $\alpha_k$, that cannot in general be estimated from data; in exchange, it replaces the unverifiable question of whether the selected concepts are informative with the measurable errors $\varepsilon_c^{(k)}, \varepsilon_s^{(k)}$. While $\Pr_{\mathcal{D}}^{(k)}(M)$ is well-defined for any concept space, estimating it from data requires selected concepts to recur across inputs. This holds for discrete or low-dimensional encodings, but not for dense continuous ones such as CLIP, so the bound is most readily evaluated in the discrete, moderate-$C$ regime. Accordingly, we treat the theory as the motivation for our faithfulness criterion rather than as a quantity measured directly in our experiments.

\subsection{Training and Inference}

The training step of \method updates only the parameters of $M$, $\widehat{M}$, and $A$, since $g$ is a pretrained model that is kept frozen. The three modules are jointly trained by optimizing a three-module game, which encourages Arthur to rely on concepts selected by Merlin, while being robust to potentially misleading concepts selected by Morgana. Given a concept encoding \(\mathbf{c} \in \mathbb{R}^C\), label \(y \in \mathcal{Y}\) and cross-entropy function $CE(\cdot, \cdot)$, we define:

\begin{itemize}
    \item \textbf{Merlin’s loss:} \quad \(L_M = CE(A(\mathcal{S}), y)\), where \(\mathcal{S}\) is the sparse concept subset selected by Merlin. This loss encourages Arthur to classify correctly based on Merlin’s input.
    
    \item \textbf{Morgana’s loss:} \quad \(L_{\widehat{M}} = CE(A(\widehat{\mathcal{S}}), y)\), where \(\widehat{\mathcal{S}} \) is Morgana’s adversarial concept subset. Here, the loss is interpreted as the classifier’s inability to be misled by deceptive inputs\footnote{In practice, the CE loss of Morgana is a slightly modified CE loss (\cf \autoref{app:rejection-loss}).}.
\end{itemize}

Overall, Arthur’s loss combines both objectives with a hyperparameter \( \gamma \in \mathbb{R}_{\geq0} \), controlling the emphasis on predictive performance (\textit{completeness}) versus robustness (\textit{soundness}): 
\begin{align}
    L_A = (1 - \gamma)\, L_M + \gamma\, L_{\widehat{M}},    
\end{align}
In detail, the three modules are updated jointly in a two-phase min-max optimization. First, the provers are updated  where Merlin minimizes \(L_M\) and Morgana maximizes \(L_{\widehat{M}}\); then Arthur is updated by minimizing \(L_A\) on the sparse selected concepts chosen by the provers. This scheme incentivizes Arthur to base its predictions on informative, task-relevant, and verifiably robust concept subsets.

At inference time, only the cooperative prover \(M\)  is used to select a sparse subset of concepts, based on the input’s concept encoding. The verifier \(A\) then predicts a label or rejects based solely on this selected subset.

Overall, by integrating concept-extractor modules and leveraging the Merlin--Arthur framework,
NCV emphasizes faithfulness and interpretability while preserving nonlinear modeling capabilities,
and shifts the min--max optimization into a lower-dimensional concept space, improving efficiency,
scalability, and stability in high-dimensional settings, a common challenge in min--max optimization
for deep learning \citep{pmlr-v80-mescheder18a, nagarajan2017gradient}. 

\section{Experimental Evaluations}
\label{sec:experimental_eval}
In this section, we present a comprehensive evaluation of \methodfull (NCV) on both synthetic and real-world high-dimensional image datasets. We evaluate based on two instantiations of \method that utilize different concept extractors: a CLIP-based extractor and the Neural Concept Binder (NCB) \citep{stammer2024neural}. We assess predictive performance and interpretability across multiple datasets, compare against several baselines and examine scalability as well as robustness against shortcut learning.

Our evaluation is structured around the following research questions: \textbf{(Q1)} Does shifting Prover-Verifier Games (PVGs) to concept-encodings via \method lead to performative classifiers (\textit{i.e.}, high completeness and soundness) on high-dimensional synthetic and real-world images? 
\textbf{(Q2)} Does \method reduce the ``interpretability-accuracy gap'' in the context of CBMs? %we further test whether NCV narrows the interpretability–accuracy gap observed in Concept Bottleneck Models (CBMs). 
\textbf{(Q3)} Does \method allow for more detailed explanations over pixel-based PVGs? %examines whether NCV provides more detailed and interpretable explanations than pixel-based PVG baselines. 
Finally, \textbf{(Q4)} Can training via \method reduce shortcut learning? %evaluates whether training within the NCV framework mitigates shortcut learning.

\subsection{Experimental Setup}
\textbf{Datasets.}
We investigate \method on CLEVR-Hans3 and CLEVR-Hans7~\citep{StammerSK21}, synthetic benchmarks derived from CLEVR~\citep{Johnson_2017_CVPR} that capture complex object compositions and include visual shortcuts. CLEVR-Hans3 features three compositional classes, while CLEVR-Hans7 increases the complexity to seven, with all images rendered at $128\!\times\!128$ pixels. The training and validation sets contain spurious correlations between attributes and labels (e.g., gray cubes linked to a specific class), which are absent in the test set, making the datasets well-suited for studying shortcut behavior. Models that exploit such correlations often fail under the decorrelated test distribution. We first report results on non-confounded versions of these datasets, where feature distributions are consistent across splits, and later return to the confounded versions for shortcut mitigation. To assess scalability and generalization to natural images, we additionally evaluate on ImageNet-1k~\citep{deng2009imagenet} with 1.2M high-resolution images across 1,000 classes (resized to $224\!\times\!224$ pixels), and on CIFAR-100~\citep{krizhevsky2009learning} with 60,000 low-resolution $32\!\times\!32$ images across 100 fine-grained categories. Lastly, we perform experiments on COCOLogic \citep{steinmann2025object}, a recent benchmark combining real-world images with complex, compositional class rules. 

\textbf{Baseline Models.} 
We compare our framework against several representative baselines, with training details provided in \autoref{app:baselines}. As a strong but non-interpretable baseline, we use a standard ResNet-18~\citep{He_2016_CVPR} for evaluations on CLEVR-Hans, and a ResNet-50 for CIFAR-100, COCOLogic and ImageNet-1k, each trained end-to-end on raw images. We further evaluate a pixel-based MAC~\citep{pmlr-v238-waldchen24a} (denoted as \textit{Pixel-MAC}), an instantiation of the Prover-Verifier Game in which the verifier is initialized from a pretrained ResNet-18, while both provers (Merlin and Morgana) are U-Net models~\citep{ronneberger2015u} that output continuous feature-importance masks over the input image. These masks are discretized using Top-$k$ selection to define the features visible to the verifier, and all agents are jointly fine-tuned; the resulting explanations (\ie, certificates) correspond to masks in pixel space (see~\citep{pmlr-v238-waldchen24a} for further details). We also compare to a vanilla Concept Bottleneck Model~\citep{koh2020concept} (denoted as \textit{CBM}), where a linear classifier predicts from concept features extracted by either NCB~\citep{stammer2024neural} for CLEVR-Hans or SpLiCE~\citep{bhalla2024interpreting} for CIFAR-100, ImageNet-1k and COCOLogic. In addition, we include a nonlinear CBM variant that replaces the linear classifier head with a nonlinear, non-interpretable head while operating on the same concept encodings, isolating the effect of a more expressive concept-level predictor.

\textbf{NCV Instantiations.}
\method is agnostic to the choice of concept extractor, and we exploit this flexibility to instantiate it across two complementary supervision paradigms: NCB for CLEVR-Hans, where its pretrained slot-based concept encodings match the object-centric structure of the synthetic dataset, and a CLIP-based extractor for the real-world datasets, where a general-purpose vocabulary derived from natural image captions is more appropriate. Within each dataset, we compare \method against baselines that share the same concept extractor, so that performance differences isolate the effect of the NCV framework rather than the choice of extractor.

For CLEVR-Hans3 and CLEVR-Hans7, we instantiate NCV with NCB~\citep{stammer2024neural} as the concept extractor, using models pretrained on CLEVR~\citep{Johnson_2017_CVPR}. A permutation-invariant Set Transformer~\citep{pmlr-v97-lee19d} serves as the verifier (Arthur) to process the unordered NCB encodings. The provers (Merlin and Morgana) are independent Set Transformers that take the full concept-slot encodings as input and output a sparse mask of 12 active concepts for the verifier. All modules are jointly trained with the Adam optimizer \citep{kingma2014adam}. Further details and ablations are provided in \autoref{app:ncb_ncv}.
For ImageNet-1k, CIFAR-100 and COCOLogic, we use a CLIP-based concept extractor~\citep{radford2021learningtransferablevisualmodels}, following the approach of SpLiCE~\citep{bhalla2024interpreting} to compute image–text similarity scores. All concepts are drawn from a fixed textual vocabulary derived from LAION captions~\citep{schuhmann2021laion} and embedded using CLIP (see \autoref{app:clip_ncv}). Unlike SpLiCE, which performs per-sample optimization, our method (denoted as \textit{CLIP-Sim}) retains the full activation vector and delegates concept selection to the provers, avoiding expensive inference-time optimization and enabling scalability. Here, the verifier and both provers are two-layer MLPs; the provers output sparse masks of 32 concepts per example. All modules are trained with the Adam optimizer. Additional details and ablations are provided in \autoref{app:ncb_ncv} and \autoref{app:clip_ncv} including the effect of varying the mask size and the weighting parameter~$\gamma$.

\begin{table*}[t!]
    \centering
    \caption{\method delivers high predictive performance and soundness through verifiable,
concept-based reasoning evaluated via completeness and soundness. We report completeness and soundness scores for ResNet, Pixel-MAC,
nonlinear and linear CBMs, and \method across synthetic (CLEVR-Hans3, CLEVR-Hans7) and real-world (CIFAR-100,
ImageNet-1k, COCOLogic) datasets. \method matches or outperforms baselines in completeness in
most settings, while offering strong
soundness guarantees.}
    \vspace{0.5em}
    \resizebox{0.98\linewidth}{!}{
    \begin{tabular}{@{}llrr rr rr@{}}
        \toprule
        Model & \makecell{Feature\\[-0.5ex]Space}
        & \makecell{Completeness\\[-0.5ex]\footnotesize(Accuracy)} & \makecell{Soundness\\[-0.5ex]\footnotesize(Robustness)} 
        & \makecell{Completeness\\[-0.5ex]\footnotesize(Accuracy)} & \makecell{Soundness\\[-0.5ex]\footnotesize(Robustness)} 
        & \makecell{Completeness\\[-0.5ex]\footnotesize(Accuracy)} & \makecell{Soundness\\[-0.5ex]\footnotesize(Robustness)} \\
        \midrule
        \multicolumn{2}{c}{} & \multicolumn{2}{c}{\textbf{CIFAR-100}} & \multicolumn{2}{c}{\textbf{ImageNet-1k}} & \multicolumn{2}{c}{\textbf{COCOLogic}} \\
        \rowcolor{gray!10} ResNet-50 & pixel space & $81.45 \std{0.60}$ & n/a & $76.01 \std{0.02}$ & n/a & $65.80 \std{3.41}$ & n/a \\
        \rowcolor{gray!10} CBM (nonlin.) & SpLiCE & $79.29 \std{0.42}$ & n/a & $69.02 \std{0.38}$ & n/a & $70.09 \std{0.56}$ & n/a \\
        \hdashline
        \addlinespace[2pt]
        Pixel-MAC  & pixel space & $15.27 \std{4.78}$ & $96.31 \std{4.12}$ & $35.06 \std{3.20}$ & $99.65 \std{0.26}$ & $42.57 \std{3.13}$ & $97.70 \std{0.61}$ \\
        CBM & SpLiCE & $75.42 \std{0.04}$ & n/a & $\boldsymbol{68.59 \std{0.01}}$ & n/a & $58.84 \std{0.09}$ & n/a \\
        \textbf{\method} (ours) & CLIP-Sim & $\boldsymbol{83.32 \std{0.28}}$ & $\boldsymbol{99.99 \std{0.01}}$ & $67.04 \std{0.16}$ & $\boldsymbol{99.94 \std{0.02}}$ & $\boldsymbol{75.42 \std{3.21}}$ & $\boldsymbol{97.87 \std{0.47}}$ \\
        \midrule
        \multicolumn{2}{c}{} & \multicolumn{2}{c}{\textbf{CLEVR-Hans3}} & \multicolumn{2}{c}{\textbf{CLEVR-Hans7}} \\
        \rowcolor{gray!10} ResNet-18 & pixel space & $97.87 \std{0.24}$ & n/a & $98.71 \std{0.24}$ & n/a \\
        \rowcolor{gray!10} CBM (nonlin.)  & NCB & $98.13 \std{0.37}$ & n/a & $97.83 \std{0.25}$ & n/a \\
        \cdashline{1-6}
        \addlinespace[2pt]
        Pixel-MAC  & pixel space & $96.59 \std{0.72}$ & $99.99 \std{0.01}$ & $97.61 \std{0.38}$ & $99.88 \std{0.28}$ \\
        CBM & NCB & $95.44 \std{0.08}$ & n/a & $89.12 \std{0.12}$ & n/a \\
        \textbf{\method} (ours)  & NCB & $\boldsymbol{98.92 \std{0.32}}$ & $\boldsymbol{100.00 \std{0.00}}$ & $\boldsymbol{97.89 \std{0.31}}$ & $\boldsymbol{100.00 \std{0.00}}$ \\
        \bottomrule
    \end{tabular}
    }
    \label{tab:scalability}
\end{table*}

\textbf{Metrics.} All methods are evaluated for completeness and, where applicable, soundness
(\cf \autoref{sec:merlin-arthur}). 
We use 20 random seeds for CLEVR-Hans and 10 for ImageNet-1k, CIFAR-100 and COCOLogic,
reporting mean and standard deviation across all seeds. For shortcut learning evaluations with CLEVR-Hans,
we additionally report a separate \emph{shortcut robustness} metric: the validation-test gap, i.e., the difference between validation accuracy on data with shortcuts and test accuracy on a split without these shortcuts.

\subsection{Evaluations}

\paragraph{Scaling PVGs to High Dimensions (Q1).}
In our first evaluation, we examine whether shifting the Prover–Verifier Game (PVG) to concept encodings enables \method to scale to high-dimensional image domains while achieving strong performance in terms of completeness and soundness. We hereby compare NCV against two key baselines: (1) a black-box ResNet classifier (ResNet-18 for CLEVR-Hans and ResNet-50 for CIFAR-100, ImageNet-1k and COCOLogic), and (2) Pixel-MAC, a nonlinear PVG model operating in raw pixel space.
\autoref{tab:scalability} summarizes results across synthetic (CLEVR-Hans3, CLEVR-Hans7) and real-world (CIFAR-100, ImageNet-1k, COCOLogic) benchmarks. Each model's feature space is indicated for clarity.
On the synthetic CLEVR-Hans benchmarks, we observe that NCV consistently achieves the highest completeness scores, surpassing Pixel-MAC and even ResNet-18 on CLEVR-Hans3, while also attaining perfect soundness. This demonstrates that NCV not only matches or exceeds the performance of strong black-box classifiers but also certifiable decision-making. Pixel-MAC performs well in these settings but falls slightly short in completeness and cannot match NCV’s zero-error soundness.

On the more challenging real-world datasets, Pixel-MAC either fails entirely or performs poorly. In contrast, NCV successfully scales to these datasets, achieving superior completeness and near-perfect soundness. Notably, NCV surpasses ResNet-50 even in raw accuracy for CIFAR-100 and COCOLogic, providing both higher predictive performance while retaining interpretability.
In summary, NCV generalizes well across domains: it scales beyond the limitations of pixel-based PVGs, delivers competitive accuracy even on large-scale and complex datasets, and retains soundness throughout. 
Overall, these findings affirm Q1: shifting PVGs to concept space yields interpretable classifiers whose decisions can be evaluated via completeness and soundness, while remaining performant and scalable on high-dimensional synthetic and real-world images.

\begin{figure*}[t!]
    \centering
    \includegraphics[width=0.88\linewidth]{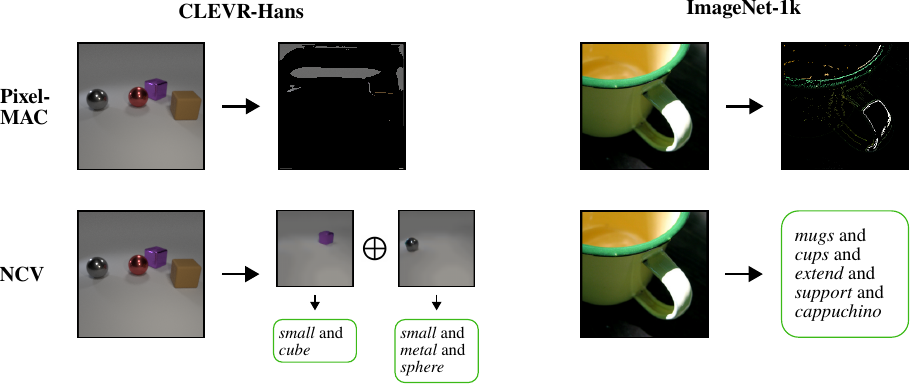}
    \caption{Comparison of explanations from \method \textit{vs.} Pixel-MAC. \textbf{(top)} Merlin–Arthur training on pixel space yields uninformative masks. \textbf{(bottom)} MAC on a concept-space via \method translates into combinations of high-level concepts and, in turn, in an interpretable prediction. 
    }
    \label{fig:qualitative_examples}
\end{figure*}

\paragraph{Narrowing the Interpretability–Accuracy Gap (Q2).}
In \autoref{tab:scalability}, we further examine whether \method can overcome a central limitation of a majority of Concept Bottleneck Models (CBMs): the interpretability–accuracy gap resulting from their use of constrained linear classifiers (\cf \autoref{app:linear} for a discussion). 
We observe that NCV consistently narrows or closes this gap while maintaining high completeness and soundness. Specifically, on CLEVR-Hans3, NCV ($\sim$99\%) exceeds the linear CBM ($\sim$95\%) and both the opaque ResNet-18 and nonlinear CBM baselines ($\sim$98\%) with perfect soundness. On CLEVR-Hans7, NCV narrows the 10-point gap left by linear CBMs to match the performance of its nonlinear counterparts.
This trend persists on real-world datasets. On CIFAR-100, NCV outperforms both the linear and nonlinear CBMs and even slightly exceeds ResNet-50’s performance. This is even more pronounced on COCOLogic, where NCV outperforms all other models by a large margin. As the additional benefits of a nonlinear classifier are quite small on ImageNet-1k, the additional training complexity of NCV results in a slightly worse performance compared to the CBMs there. 
Overall, NCV improves over linear CBMs in accuracy, especially on tasks requiring complex concept reasoning such as COCOLogic. At the same time, NCV can match or even surpass the accuracy of opaque ResNet models, demonstrating that interpretable, concept-level classification via Prover–Verifier Games can deliver competitive performance. Additional comparisons with Deep Concept Reasoning (DCR; \citealt{barbiero2023interpretable}) using the same concept vocabulary are reported in the supplementary material, highlighting that \method can also outperform CBMs with nonlinear, yet interpretable classifiers on large-scale datasets (\cf \autoref{tab:dcr_comparison} in \autoref{app:baselines}). We therefore answer Q2 affirmatively. 

\paragraph{More Detailed Explanations (Q3)} 
We next investigate the resulting explanations produced by \method, with a focus on explanatory clarity. Since our goal is to improve over classic vision-based Prover–Verifier Games, we compare against pixel-level MAC explanations. \autoref{fig:qualitative_examples} illustrates a qualitative example from both the CLEVR‐Hans3 (\cf \autoref{fig:CLEVRHans3_explanations_allclasses} for more examples) and ImageNet-1k datasets. 
Notably, under Pixel‐MAC, the Prover–Verifier setup operates directly on pixels, yielding broad, diffuse explanation masks that often cover entire objects or irrelevant background regions, arguably providing limited insight regarding which exact features drive the verifier's final decision. 
In contrast, NCV leverages its internal concept encodings to isolate sparse, high-level concepts that are consistently associated with a class decision under the prover--verifier interaction, recovering the class rule for CLEVR-Hans (i.e., small cube and small metal sphere) and providing a meaningful concept explanation for the coffee-mug class of ImageNet-1k.\footnote{The availability of object-level concepts in \method\ depends on the underlying concept extractor. For CLEVR-Hans, we use NCB, which provides such object-based explanations.} For clarity, we only show the five most frequent concepts (from a total mask size of 32) across 32 samples for ImageNet-1k. All concepts are part of the CLIP/SpLiCE-based concept vocabulary that is derived from LAION captions~\citep{schuhmann2021laion} (see \autoref{app:clip_ncv}; \autoref{app:concept-comparison} compares them against a linear CBM).

Overall, these examples highlight that NCV offers higher-level, semantically meaningful explanations rather than fine-grained pixel masks, and that concept-level PVGs yield interpretable decisions whose supporting concept subsets can be evaluated even for complex, high-dimensional data.
This leads us to answer Q3 affirmatively.

\begin{table*}[t!]
    \caption{Shortcut robustness on CLEVR-Hans3 and CLEVR-Hans7. We report validation
accuracy on a shortcut-confounded split, test accuracy on a clean split, and the resulting
validation--test gap (``shortcut robustness'', lower is better) across models trained with varying
amounts of clean data. When any amount of clean data is available, \method shows improvements compared to training without PVGs.}
    \label{tab:rq4}

    \centering
    \resizebox{0.85\textwidth}{!}{%
    \begin{tabular}{cl|ccc|ccc}
    \toprule
    \multirow{2}{*}{\makecell[c]{Ratio\\Non-Conf.\\{\scriptsize(Samples)}}} 
    & \multirow{2}{*}{\makecell[c]{Model}} 
    & \multicolumn{3}{c|}{CLEVR-Hans3} 
    & \multicolumn{3}{c}{CLEVR-Hans7} \\
    & 
    & \makecell{Val Acc\\{\scriptsize(w/ shortcut)}} 
    & \makecell{Test Acc\\{\scriptsize(w/o shortcut)}} 
    & \makecell{Val-Test\\Gap (\textdownarrow)} 
    & \makecell{Val Acc\\{\scriptsize(w/ shortcut)}} 
    & \makecell{Test Acc\\{\scriptsize(w/o shortcut)}} 
    & \makecell{Val-Test\\Gap (\textdownarrow)} \\
    \midrule

    \multirow{3}{*}{0\%} 
     & CBM (lin.) & $95.65 \std{0.09}$ & $90.54 \std{0.09}$ & $5.11$ 
                  & $90.37 \std{0.10}$ & $85.27 \std{0.15}$ & $\mathbf{5.10}$ \\
     & CBM (non-lin.) & $98.70 \std{0.32}$ & $\mathbf{95.04} \std{0.96}$ & $\mathbf{3.66}$ 
                      & $98.09 \std{0.24}$ & $90.69 \std{1.17}$ & $7.40$ \\
     & \textbf{\method} & $\mathbf{99.44} \std{0.15}$ & $94.21 \std{1.41}$ & $5.23$ 
                & $\mathbf{98.38} \std{0.18}$ & $\mathbf{92.23} \std{0.67}$ & $6.15$ \\
    \midrule    

    \multirow{3}{*}{\makecell{1\%\\(105)}} 
     & CBM (lin.) & $96.28 \std{0.16}$ & $91.03 \std{0.31}$ & $5.25$ 
                  & $90.74 \std{0.12}$ & $85.41 \std{0.17}$ & $5.33$ \\
     & CBM (non-lin.) & $99.10 \std{0.27}$ & $94.84 \std{0.98}$ & $4.26$ 
                      & $98.17 \std{0.17}$ & $92.65 \std{1.31}$ & $5.52$ \\
     & \textbf{\method} & $\mathbf{99.37} \std{0.18}$ & $\mathbf{97.11} \std{0.98}$ & $\mathbf{2.26}$ 
                & $\mathbf{98.19} \std{0.24}$ & $\mathbf{94.68} \std{0.64}$ & $\mathbf{3.51}$ \\
    \midrule    

    \multirow{3}{*}{\makecell{5\%\\(525)}} 
     & CBM (lin.) & $95.38 \std{0.37}$ & $93.34 \std{0.51}$ & $2.04$ 
                  & $90.37 \std{0.15}$ & $86.37 \std{0.18}$ & $4.00$ \\
     & CBM (non-lin.) & $98.41 \std{0.55}$ & $96.13 \std{0.71}$ & $2.28$ 
                      & $98.32 \std{0.22}$ & $95.19 \std{0.80}$ & $3.13$ \\
     & \textbf{\method} & $\mathbf{99.59} \std{0.19}$ & $\mathbf{98.88} \std{0.37}$ & $\mathbf{0.71}$ 
                & $\mathbf{98.47} \std{0.24}$ & $\mathbf{96.24} \std{0.71}$ & $\mathbf{2.23}$ \\
    \midrule     

    \multirow{3}{*}{\makecell{20\%\\(2100)}} 
     & CBM (lin.) & $95.67 \std{0.28}$ & $93.46 \std{0.23}$ & $2.21$ 
                  & $89.93 \std{0.29}$ & $87.21 \std{0.31}$ & $2.72$ \\
     & CBM (non-lin.) & $99.15 \std{0.21}$ & $98.09 \std{0.51}$ & $1.06$ 
                      & $98.21 \std{0.29}$ & $97.00 \std{0.49}$ & $1.21$ \\
     & \textbf{\method} & $\mathbf{99.37} \std{0.28}$ & $\mathbf{98.82} \std{0.67}$ & $\mathbf{0.55}$ 
                & $\mathbf{98.63} \std{0.13}$ & $\mathbf{97.74} \std{0.28}$ & $\mathbf{0.89}$ \\    
    \bottomrule
    \end{tabular}%
    }
\end{table*}

\paragraph{Mitigating Shortcut Learning (Q4)}
Lastly, to assess whether \method can mitigate shortcut learning in image classification, we train models on different versions of CLEVR-Hans3 and CLEVR-Hans7 with varying ratios of clean samples (\textit{i.e.}, without shortcut) in the training and validation sets. We then measure validation accuracy with shortcuts and test accuracy on a held-out, clean data split. This setup allows us to track both predictive performance and robustness to shortcut learning. \autoref{tab:rq4} reports results for three model types: a linear CBM, a nonlinear CBM, and our instantiation of \method using NCB as concept extractor. We observe that while \method achieves the highest test accuracy among all models in the 0\% clean data setting, it still exhibits a sizeable validation-test gap, indicating a strong influence of the underlying shortcuts. %Notably, \method shows a smaller gap than the nonlinear CBM, suggesting that the inductive bias of the PVG-based training objective encourages a smaller influence of the shortcut, even without access to clean training data.
As the amount of clean samples is progressively increased, test accuracy and test-validation gap improves across all models. However, \method consistently achieves the highest test accuracy in every setting, and its validation–test gap decreases more rapidly than for either CBM variant. This trend indicates that \method is not only better at leveraging clean supervision when available, but also exhibits a faster reduction of shortcut reliance as clean data increases. %A similar trend is observed for CLEVR-Hans3 (\cf \autoref{tab:rq4_clevrhans3}), where \method closes the generalization gap even more efficiently than all baselines. 
Together, these results demonstrate that concept-level prover–verifier interaction in \method encourage models to rely on more robust, task-relevant features, making \method more resilient to shortcut learning, even with limited amounts of clean data.

\section{Discussion}\label{sec:discussion}
Overall, our results show that shifting Prover–Verifier Games (PVGs) to the concept level yields a powerful and scalable framework for verifiable, interpretable classification. By operating on symbolic concept embeddings, \method avoids the computational cost of per-sample inference in pixel space, yet matches or surpasses pixel-based baselines in both completeness and soundness. It reduces the performance gap typical of Concept Bottleneck Models (CBMs), achieving parity with opaque models on synthetic tasks and even surpassing them on natural images. Concept-level outputs offer concise, human-readable explanations. Finally, NCV exhibits higher resilience to spurious correlations, generalizing from confounded training splits and closing the generalization gap with minimal available clean data.

That said, \method has several limitations. Its effectiveness depends on the quality of the underlying concept extractor: noisy or entangled concept spaces can reduce both accuracy and human understandability. The increased training complexity introduced by the three-module PVG setup also results in greater computational cost and training instability, e.g., compared to linear CBMs. 
Moreover, when using pretrained models like CLIP for concept discovery, \method inherits their biases and inconsistencies to some extent \citep{birhane2021multimodal, gehman2020realtoxicityprompts, bhalla2024interpreting}. Finally, recent work~\citep{debole2025if} shows that such concept spaces can diverge from expert semantics, even when yielding strong downstream performance. 

\section{Conclusion}
\label{sec:conclusion}
In this work, we have introduced the Neural Concept Verifier (NCV), a unified framework that combines Prover–Verifier Games (PVGs) and concept‐level representations for interpretable classification at scale. Through extensive experiments on CLEVR-Hans, CIFAR-100, ImageNet-1k, and COCOLogic, we have shown that NCV scales well to high-dimensional datasets by achieving high completeness and soundness, reduces the interpretability–accuracy gap of concept bottleneck models, delivers detailed concept-based explanations, and helps mitigate shortcut learning when clean supervision is available. Thus, NCV provides a promising pathway for deploying trustworthy and transparent models in domains where both predictive performance and verifiabile explanations are essential.

Future work should explore how concept encodings can be integrated into alternative PVG-style setups, where structured representations may improve performance or reduce communication overhead. It is also promising to investigate applications such as natural language processing and structured data, where interpretable verification may be equally valuable. Finally, at the optimization level, developing more stable schemes for training Merlin and Morgana end-to-end on discrete, binarized masks could further strengthen both the framework and the provers themselves.

\clearpage

\section*{Impact Statement}
This work aims to advance methods for interpretable and verifiable machine learning models. While improved interpretability may support safer deployment in high-stakes settings, we do not foresee any negative societal impacts beyond those generally associated with machine learning research.

\section*{Acknowledgements}
This research was partially supported by the Deutsche Forschungsgemeinschaft (DFG, German Research Foundation) under Germany's Excellence Strategy -- The Berlin Mathematics Research Center MATH+ (EXC-2046/1, EXC-2046/2, project ID: 390685689). This work was further supported by the DFG under Germany’s Excellence Strategy (EXC 3066/1 “The Adaptive Mind”, Project No. 533717223) and by the ``ML2MT'' project from the Volkswagen Stiftung. It has further benefited from the HMWK projects ``The Third Wave of Artificial Intelligence - 3AI'', and Hessian.AI, the EU-funded ``TANGO'' project (EU Horizon 2023, GA No 57100431), from early stages of the Cluster of Excellence ``Reasonable AI'' funded by the DFG under Germany's Excellence Strategy -- EXC-3057 and the research collaboration between TU Darmstadt and Aleph Alpha Research through Lab1141. It has also benefited from the DFG project GRK 2853 ``Neuroexplicit Models of Language, Vision, and Action'' (project number 471607914).

\section*{Software and Data}
The code and data used in this work are publicly available at
\url{https://github.com/ZIB-IOL/Neural-Concept-Verifier} to facilitate reproducibility.
% Authors are \textbf{required} to include a statement of the potential broader
% impact of their work, including its ethical aspects and future societal
% consequences. This statement should be in an unnumbered section at the end of
% the paper (co-located with Acknowledgements -- the two may appear in either
% order, but both must be before References), and does not count toward the paper
% page limit. In many cases, where the ethical impacts and expected societal
% implications are those that are well established when advancing the field of
% Machine Learning, substantial discussion is not required, and a simple
% statement such as the following will suffice:

% ``This paper presents work whose goal is to advance the field of Machine
% Learning. There are many potential societal consequences of our work, none
% which we feel must be specifically highlighted here.''

% The above statement can be used verbatim in such cases, but we encourage
% authors to think about whether there is content which does warrant further
% discussion, as this statement will be apparent if the paper is later flagged
% for ethics review.

\bibliography{sample}

@article{adebayo2018sanity,
  title={Sanity checks for saliency maps},
  author={Adebayo, Julius and Gilmer, Justin and Muelly, Michael and Goodfellow, Ian and Hardt, Moritz and Kim, Been},
  journal={Advances in neural information processing systems},
  volume={31},
  year={2018}
}

@article{de2025causally,
  author  = {De Felice, Giovanni and Flores, Arianna Casanova and De Santis, Francesco and Santini, Silvia and Schneider, Johannes and Barbiero, Pietro and Termine, Alberto},
  title   = {Causally reliable concept bottleneck models},
  journal = {arXiv preprint arXiv:2503.04363},
  year    = {2025}
}

@article{dominici2024causal,
  author  = {Dominici, Gabriele and Barbiero, Pietro and Zarlenga, Mateo Espinosa and Termine, Alberto and Gjoreski, Martin and Marra, Giuseppe and Langheinrich, Marc},
  title   = {Causal concept graph models: Beyond causal opacity in deep learning},
  journal = {arXiv preprint arXiv:2405.16507},
  year    = {2024}
}

@article{debot2024interpretable,
  author  = {Debot, David and Barbiero, Pietro and Giannini, Francesco and Ciravegna, Gabriele and Diligenti, Michelangelo and Marra, Giuseppe},
  title   = {Interpretable concept-based memory reasoning},
  journal = {Advances in Neural Information Processing Systems},
  volume  = {37},
  pages   = {19254--19287},
  year    = {2024}
}

@inproceedings{barbiero2023interpretable,
  author       = {Barbiero, Pietro and Ciravegna, Gabriele and Giannini, Francesco and Zarlenga, Mateo Espinosa and Magister, Lucie Charlotte and Tonda, Alberto and Li{\'o}, Pietro and Precioso, Frederic and Jamnik, Mateja and Marra, Giuseppe},
  title        = {Interpretable neural-symbolic concept reasoning},
  booktitle    = {International Conference on Machine Learning},
  pages        = {1801--1825},
  year         = {2023},
  organization = {PMLR}
}

@inproceedings{espinosa2022concept,
  author    = {Zarlenga, Mateo Espinosa and Barbiero, Pietro and Ciravegna, Gabriele and Marra, Giuseppe and Giannini, Francesco and Diligenti, Michelangelo and Shams, Zohreh and Precioso, Frederic and Melacci, Stefano and Weller, Adrian and Lio, Pietro and Jamnik, Mateja},
  title     = {Concept embedding models: beyond the accuracy-explainability trade-off},
  booktitle = {Proceedings of the 36th International Conference on Neural Information Processing Systems},
  year      = {2022},
  isbn      = {9781713871088},
  publisher = {Curran Associates Inc.},
  address   = {Red Hook, NY, USA},
  articleno = {1555},
  numpages  = {14},
  location  = {New Orleans, LA, USA},
  series    = {NIPS '22}
}

@inproceedings{koh2020concept,
  author       = {Koh, Pang Wei and Nguyen, Thao and Tang, Yew Siang and Mussmann, Stephen and Pierson, Emma and Kim, Been and Liang, Percy},
  title        = {Concept bottleneck models},
  booktitle    = {International Conference on Machine Learning},
  pages        = {5338--5348},
  year         = {2020},
  organization = {PMLR}
}

@article{oikarinen2023label,
  author  = {Oikarinen, Tuomas and Das, Subhro and Nguyen, Lam M. and Weng, Tsui-Wei},
  title   = {Label-free concept bottleneck models},
  journal = {arXiv preprint arXiv:2304.06129},
  year    = {2023}
}

@inproceedings{panousis2024coarse,
  author    = {Panousis, Konstantinos P. and Ienco, Dino and Marcos, Diego},
  title     = {Coarse-to-fine concept bottleneck models},
  booktitle = {Proceedings of the 38th International Conference on Neural Information Processing Systems},
  year      = {2024},
  isbn      = {9798331314385},
  publisher = {Curran Associates Inc.},
  address   = {Red Hook, NY, USA},
  articleno = {3340},
  numpages  = {29},
  location  = {Vancouver, BC, Canada},
  series    = {NIPS '24}
}

@article{debole2025if,
  author  = {Debole, Nicola and Barbiero, Pietro and Giannini, Francesco and Passeggini, Andrea and Teso, Stefano and Marconato, Emanuele},
  title   = {If Concept Bottlenecks are the Question, are Foundation Models the Answer?},
  journal = {arXiv preprint arXiv:2504.19774},
  year    = {2025}
}

@article{steinmann2025object,
  author  = {Steinmann, David and Stammer, Wolfgang and W{\"u}st, Antonia and Kersting, Kristian},
  title   = {Object Centric Concept Bottlenecks},
  journal = {arXiv preprint arXiv:2505.24492},
  year    = {2025}
}

@inproceedings{stammer2024neural,
  author    = {Stammer, Wolfgang and W{\"u}st, Antonia and Steinmann, David and Kersting, Kristian},
  title     = {Neural Concept Binder},
  booktitle = {Advances in Neural Information Processing Systems},
  pages     = {71792--71830},
  year      = {2024},
  publisher = {Curran Associates, Inc.}
}

@article{bhalla2024interpreting,
  author  = {Bhalla, Usha and Oesterling, Alex and Srinivas, Suraj and Calmon, Flavio and Lakkaraju, Himabindu},
  title   = {Interpreting {CLIP} with sparse linear concept embeddings ({SPLICE})},
  journal = {Advances in Neural Information Processing Systems},
  volume  = {37},
  pages   = {84298--84328},
  year    = {2024}
}

@inproceedings{yang2023language,
  author    = {Yang, Yue and Panagopoulou, Artemis and Zhou, Shenghao and Jin, Daniel and Callison-Burch, Chris and Yatskar, Mark},
  title     = {Language in a bottle: Language model guided concept bottlenecks for interpretable image classification},
  booktitle = {IEEE/CVF Conference on Computer Vision and Pattern Recognition},
  pages     = {19187--19197},
  year      = {2023}
}

@article{rudin2019stop,
  author  = {Rudin, Cynthia},
  title   = {Stop explaining black box models for high stakes decisions and use interpretable models instead},
  journal = {Nature Machine Intelligence},
  volume  = {1},
  pages   = {206--215},
  year    = {2019}
}

@article{mahinpei2021promises,
  author  = {Mahinpei, Anita and Clark, Justin and Lage, Isaac and Doshi-Velez, Finale and Pan, Weiwei},
  title   = {Promises and pitfalls of black-box concept learning models},
  journal = {arXiv preprint arXiv:2106.13314},
  year    = {2021}
}

@inproceedings{pmlr-v238-waldchen24a,
  author    = {W\"{a}ldchen, Stephan and Sharma, Kartikey and Turan, Berkant and Zimmer, Max and Pokutta, Sebastian},
  title     = {Interpretability Guarantees with {M}erlin-{A}rthur Classifiers},
  booktitle = {International Conference on Artificial Intelligence and Statistics},
  pages     = {1963--1971},
  year      = {2024},
  volume    = {238},
  publisher = {PMLR}
}

@article{ghorbani2019towards,
  author  = {Ghorbani, Amirata and Wexler, James and Zou, James Y. and Kim, Been},
  title   = {Towards automatic concept-based explanations},
  journal = {Advances in Neural Information Processing Systems},
  year    = {2019}
}

@article{schut2025bridging,
  author    = {Schut, Lisa and Toma{\v{s}}ev, Nenad and McGrath, Thomas and Hassabis, Demis and Paquet, Ulrich and Kim, Been},
  title     = {Bridging the human--{AI} knowledge gap through concept discovery and transfer in {AlphaZero}},
  journal   = {Proceedings of the National Academy of Sciences},
  volume    = {122},
  number    = {13},
  year      = {2025},
  publisher = {National Academy of Sciences}
}

@inproceedings{kim2018interpretability,
  author       = {Kim, Been and Wattenberg, Martin and Gilmer, Justin and Cai, Carrie and Wexler, James and Viegas, Fernanda},
  title        = {Interpretability beyond feature attribution: Quantitative testing with concept activation vectors ({TCAV})},
  booktitle    = {International Conference on Machine Learning},
  pages        = {2668--2677},
  year         = {2018},
  organization = {PMLR}
}

@article{crabbe2022concept,
  author  = {Crabb{\'e}, Jonathan and van der Schaar, Mihaela},
  title   = {Concept activation regions: A generalized framework for concept-based explanations},
  journal = {Advances in Neural Information Processing Systems},
  pages   = {2590--2607},
  year    = {2022}
}

@article{lee2025neural,
  author    = {Lee, Jae Hee and Lanza, Sergio and Wermter, Stefan},
  title     = {From neural activations to concepts: A survey on explaining concepts in neural networks},
  journal   = {Neurosymbolic Artificial Intelligence},
  volume    = {1},
  pages     = {NAI--240743},
  year      = {2025},
  publisher = {SAGE Publications}
}

@article{poeta2023concept,
  author  = {Poeta, Eleonora and Ciravegna, Gabriele and Pastor, Eliana and Cerquitelli, Tania and Baralis, Elena},
  title   = {Concept-based explainable artificial intelligence: A survey},
  journal = {arXiv preprint arXiv:2312.12936},
  year    = {2023}
}

@article{de2025towards,
  author  = {De Santis, Francesco and Bich, Philippe and Ciravegna, Gabriele and Barbiero, Pietro and Giordano, Danilo and Cerquitelli, Tania},
  title   = {Towards Better Generalization and Interpretability in Unsupervised Concept-Based Models},
  journal = {arXiv preprint arXiv:2506.02092},
  year    = {2025}
}

@inproceedings{browncohen2024scalable,
  author    = {Brown-Cohen, Jonah and Irving, Geoffrey and Piliouras, Georgios},
  title     = {Scalable {AI} safety via doubly-efficient debate},
  booktitle = {International Conference on Machine Learning},
  year      = {2024},
  publisher = {JMLR.org}
}

@inproceedings{goldwasserproofs,
  author    = {Goldwasser, Shafi and Micali, Silvio and Rackoff, Charles},
  title     = {The knowledge complexity of interactive proof-systems},
  booktitle = {ACM Symposium on Theory of Computing},
  pages     = {291--304},
  year      = {1985},
  publisher = {ACM}
}

@article{anil2021learning,
  author  = {Anil, Cem and Zhang, Guodong and Wu, Yuhuai and Grosse, Roger},
  title   = {Learning to give checkable answers with prover-verifier games},
  journal = {arXiv preprint arXiv:2108.12099},
  year    = {2021}
}

@article{amit2024models,
  author  = {Amit, Noga and Goldwasser, Shafi and Paradise, Orr and Rothblum, Guy},
  title   = {Models that prove their own correctness},
  journal = {arXiv preprint arXiv:2405.15722},
  year    = {2024}
}

@article{irving2018aisafetydebate,
  author  = {Irving, Geoffrey and Christiano, Paul and Amodei, Dario},
  title   = {{AI} safety via debate},
  journal = {arXiv preprint arXiv:1805.00899},
  year    = {2018}
}

@article{Kirchner24PV,
  author  = {Kirchner, Jan Hendrik and Chen, Yining and Edwards, Harri and Leike, Jan and McAleese, Nat and Burda, Yuri},
  title   = {Prover-verifier games improve legibility of {LLM} outputs},
  journal = {arXiv preprint arXiv:2407.13692},
  year    = {2024}
}

@inproceedings{Du23multiagent,
  author    = {Du, Yilun and Li, Shuang and Torralba, Antonio and Tenenbaum, Joshua B. and Mordatch, Igor},
  title     = {Improving Factuality and Reasoning in Language Models through Multiagent Debate},
  booktitle = {International Conference on Machine Learning},
  pages     = {11733--11763},
  year      = {2024},
  volume    = {235},
  publisher = {PMLR}
}

@article{SchramowskiSTBH20,
  author  = {Schramowski, Patrick and Stammer, Wolfgang and Teso, Stefano and Brugger, Anna and Herbert, Franziska and Shao, Xiaoting and Luigs, Hans-Georg and Mahlein, Anne-Katrin and Kersting, Kristian},
  title   = {Making deep neural networks right for the right scientific reasons by interacting with their explanations},
  journal = {Nature Machine Intelligence},
  volume  = {2},
  number  = {8},
  pages   = {476--486},
  year    = {2020}
}

@article{FriedrichSSK23b,
  author  = {Friedrich, Felix and Stammer, Wolfgang and Schramowski, Patrick and Kersting, Kristian},
  title   = {A typology for exploring the mitigation of shortcut behaviour},
  journal = {Nature Machine Intelligence},
  volume  = {5},
  number  = {3},
  pages   = {319--330},
  year    = {2023}
}

@article{steinmann2024navigating,
  author  = {Steinmann, David and Divo, Felix and Kraus, Maurice and W{\"u}st, Antonia and Struppek, Lukas and Friedrich, Felix and Kersting, Kristian},
  title   = {Navigating Shortcuts, Spurious Correlations, and Confounders: From Origins via Detection to Mitigation},
  journal = {arXiv preprint arXiv:2412.05152},
  year    = {2024}
}

@article{geirhos2020shortcut,
  author    = {Geirhos, Robert and Jacobsen, J{\"o}rn-Henrik and Michaelis, Claudio and Zemel, Richard and Brendel, Wieland and Bethge, Matthias and Wichmann, Felix A.},
  title     = {Shortcut learning in deep neural networks},
  journal   = {Nature Machine Intelligence},
  volume    = {2},
  number    = {11},
  pages     = {665--673},
  year      = {2020},
  publisher = {Nature Publishing Group UK London}
}

@article{Bortolotti25nesy,
  author  = {Bortolotti, Samuele and Marconato, Emanuele and Morettin, Paolo and Passerini, Andrea and Teso, Stefano},
  title   = {Shortcuts and Identifiability in Concept-based Models from a Neuro-Symbolic Lens},
  journal = {arXiv preprint arXiv:2502.11245},
  year    = {2025}
}

@article{MarconatoTVP23,
  author  = {Marconato, Emanuele and Teso, Stefano and Vergari, Antonio and Passerini, Andrea},
  title   = {Not all neuro-symbolic concepts are created equal: Analysis and mitigation of reasoning shortcuts},
  journal = {Advances in Neural Information Processing Systems},
  volume  = {36},
  pages   = {72507--72539},
  year    = {2023}
}

@inproceedings{ross2017right,
  author    = {Ross, Andrew Slavin and Hughes, Michael C. and Doshi-Velez, Finale},
  title     = {Right for the Right Reasons: Training Differentiable Models by Constraining their Explanations},
  booktitle = {International Joint Conference on Artificial Intelligence},
  pages     = {2662--2670},
  year      = {2017}
}

@inproceedings{StammerSK21,
  author    = {Stammer, Wolfgang and Schramowski, Patrick and Kersting, Kristian},
  title     = {Right for the Right Concept: Revising Neuro-Symbolic Concepts by Interacting With Their Explanations},
  booktitle = {IEEE/CVF Conference on Computer Vision and Pattern Recognition},
  pages     = {3619--3629},
  year      = {2021}
}

@article{lapuschkin2019unmasking,
  author    = {Lapuschkin, Sebastian and W{\"a}ldchen, Stephan and Binder, Alexander and Montavon, Gr{\'e}goire and Samek, Wojciech and M{\"u}ller, Klaus-Robert},
  title     = {Unmasking Clever Hans predictors and assessing what machines really learn},
  journal   = {Nature Communications},
  volume    = {10},
  number    = {1},
  pages     = {1--8},
  year      = {2019},
  publisher = {Nature Publishing Group}
}

@article{ahmed2021discovery,
  author    = {Ahmed, Kaoutar Ben and Goldgof, Gregory M. and Paul, Rahul and Goldgof, Dmitry B. and Hall, Lawrence O.},
  title     = {Discovery of a generalization gap of convolutional neural networks on {COVID-19} X-rays classification},
  journal   = {IEEE Access},
  volume    = {9},
  pages     = {72970--72979},
  year      = {2021},
  publisher = {IEEE}
}

@article{fok23,
  author    = {Fok, Raymond and Weld, Daniel S.},
  title     = {In search of verifiability: Explanations rarely enable complementary performance in {AI}-advised decision making},
  journal   = {AI Magazine},
  year      = {2023},
  publisher = {Wiley Online Library}
}

@article{Nair23,
  author  = {Nair, Varun and Schumacher, Elliot and Tso, Geoffrey and Kannan, Anitha},
  title   = {{DERA}: enhancing large language model completions with dialog-enabled resolving agents},
  journal = {arXiv preprint arXiv:2303.17071},
  year    = {2023}
}

@article{PruthiBDSCLNC22,
  title     = {Evaluating Explanations: How Much Do Explanations from the Teacher Aid Students?},
  author    = {Pruthi, Danish and Bansal, Rachit and Dhingra, Bhuwan and Baldini Soares, Livio and Collins, Michael and Lipton, Zachary C. and Neubig, Graham and Cohen, William W.},
  editor    = {Roark, Brian and Nenkova, Ani},
  journal   = {Transactions of the Association for Computational Linguistics},
  volume    = {10},
  year      = {2022},
  address   = {Cambridge, MA},
  publisher = {MIT Press},
  url       = {https://aclanthology.org/2022.tacl-1.21/},
  doi       = {10.1162/tacl_a_00465},
  pages     = {359--375}
}

@article{schneider2023reflective,
  author    = {Schneider, Johannes and Vlachos, Michalis},
  title     = {Reflective-net: Learning from explanations},
  journal   = {Data Mining and Knowledge Discovery},
  volume    = {38},
  number    = {5},
  pages     = {2975--2996},
  year      = {2024},
  publisher = {Springer}
}

@article{stammer2024learning,
  author  = {Stammer, Wolfgang and Friedrich, Felix and Steinmann, David and Brack, Manuel and Shindo, Hikaru and Kersting, Kristian},
  title   = {Learning by Self-Explaining},
  journal = {Transactions on Machine Learning Research},
  year    = {2024}
}

@article{kingma2014adam,
  author  = {Kingma, Diederik P. and Ba, Jimmy},
  title   = {Adam: A method for stochastic optimization},
  journal = {arXiv preprint arXiv:1412.6980},
  year    = {2014}
}

@article{paszke2019pytorch,
  author  = {Paszke, Adam and Gross, Sam and Massa, Francisco and Lerer, Adam and Bradbury, James and Chanan, Gregory and Killeen, Trevor and Lin, Zeming and Gimelshein, Natalia and Antiga, Luca},
  title   = {{PyTorch}: An imperative style, high-performance deep learning library},
  journal = {Advances in Neural Information Processing Systems},
  volume  = {32},
  year    = {2019}
}

@article{hendrycks2016gaussian,
  author  = {Hendrycks, Dan and Gimpel, Kevin},
  title   = {Gaussian error linear units ({GELUs})},
  journal = {arXiv preprint arXiv:1606.08415},
  year    = {2016}
}

@inproceedings{ronneberger2015u,
  author       = {Ronneberger, Olaf and Fischer, Philipp and Brox, Thomas},
  title        = {U-net: Convolutional networks for biomedical image segmentation},
  booktitle    = {Medical Image Computing and Computer-Assisted Intervention},
  pages        = {234--241},
  year         = {2015},
  organization = {Springer}
}

@inproceedings{He_2016_CVPR,
  author    = {He, Kaiming and Zhang, Xiangyu and Ren, Shaoqing and Sun, Jian},
  title     = {Deep Residual Learning for Image Recognition},
  booktitle = {IEEE Conference on Computer Vision and Pattern Recognition},
  year      = {2016}
}

@InProceedings{pmlr-v97-lee19d,
  title     = {Set Transformer: A Framework for Attention-based Permutation-Invariant Neural Networks},
  author    = {Lee, Juho and Lee, Yoonho and Kim, Jungtaek and Kosiorek, Adam and Choi, Seungjin and Teh, Yee Whye},
  booktitle = {Proceedings of the 36th International Conference on Machine Learning},
  pages     = {3744--3753},
  year      = {2019},
  editor    = {Chaudhuri, Kamalika and Salakhutdinov, Ruslan},
  volume    = {97},
  series    = {Proceedings of Machine Learning Research},
  month     = {09--15 Jun},
  publisher = {PMLR},
  url       = {https://proceedings.mlr.press/v97/lee19d.html}
}

@article{kimura2024permutation,
  author  = {Kimura, Masanari and Shimizu, Ryotaro and Hirakawa, Yuki and Goto, Ryosuke and Saito, Yuki},
  title   = {On permutation-invariant neural networks},
  journal = {arXiv preprint arXiv:2403.17410},
  year    = {2024}
}

@article{gluch2024goodbaduglywatermarks,
  author  = {G{\l}uch, Grzegorz and Turan, Berkant and Nagarajan, Sai Ganesh and Pokutta, Sebastian},
  title   = {The {G}ood, the {B}ad and the {U}gly: Meta-Analysis of Watermarks, {T}ransferable {A}ttacks and {A}dversarial {D}efenses},
  journal = {arXiv preprint arXiv:2410.08864},
  year    = {2024}
}

@inproceedings{pmlr-v80-mescheder18a,
  author    = {Mescheder, Lars and Geiger, Andreas and Nowozin, Sebastian},
  title     = {Which Training Methods for {GAN}s do actually Converge?},
  booktitle = {International Conference on Machine Learning},
  pages     = {3481--3490},
  year      = {2018},
  publisher = {PMLR}
}

@inproceedings{nagarajan2017gradient,
  author    = {Nagarajan, Vaishnavh and Kolter, J. Zico},
  title     = {Gradient descent {GAN} optimization is locally stable},
  booktitle = {Advances in Neural Information Processing Systems},
  pages     = {5591--5600},
  year      = {2017},
  publisher = {Curran Associates Inc.}
}

@article{deng2009imagenet,
  author       = {Deng, Jia and Dong, Wei and Socher, Richard and Li, Li-Jia and Li, Kai and Fei-Fei, Li},
  title        = {{ImageNet}: A large-scale hierarchical image database},
  journal      = {IEEE Conference on Computer Vision and Pattern Recognition},
  pages        = {248--255},
  year         = {2009},
  organization = {IEEE}
}

@inproceedings{radford2021learningtransferablevisualmodels,
  author       = {Radford, Alec and Kim, Jong Wook and Hallacy, Chris and Ramesh, Aditya and Goh, Gabriel and Agarwal, Sandhini and Sastry, Girish and Askell, Amanda and Mishkin, Pamela and Clark, Jack},
  title        = {Learning transferable visual models from natural language supervision},
  booktitle    = {International Conference on Machine Learning},
  pages        = {8748--8763},
  year         = {2021},
  organization = {PMLR}
}

@techreport{krizhevsky2009learning,
  author      = {Krizhevsky, Alex},
  title       = {Learning multiple layers of features from tiny images},
  year        = {2009},
  institution = {University of Toronto}
}

@inproceedings{Johnson_2017_CVPR,
  author    = {Johnson, Justin and Hariharan, Bharath and van der Maaten, Laurens and Fei-Fei, Li and Lawrence Zitnick, C. and Girshick, Ross},
  title     = {{CLEVR}: A Diagnostic Dataset for Compositional Language and Elementary Visual Reasoning},
  booktitle = {IEEE Conference on Computer Vision and Pattern Recognition},
  year      = {2017}
}

@article{schuhmann2021laion,
  author  = {Schuhmann, Christoph and Vencu, Richard and Beaumont, Romain and Kaczmarczyk, Robert and Mullis, Clayton and Katta, Aarush and Coombes, Theo and Jitsev, Jenia and Komatsuzaki, Aran},
  title   = {{LAION-400M}: Open dataset of {CLIP}-filtered 400 million image-text pairs},
  journal = {arXiv preprint arXiv:2111.02114},
  year    = {2021}
}

@article{gehman2020realtoxicityprompts,
  author  = {Gehman, Samuel and Gururangan, Suchin and Sap, Maarten and Choi, Yejin and Smith, Noah A.},
  title   = {{RealToxicityPrompts}: Evaluating neural toxic degeneration in language models},
  journal = {arXiv preprint arXiv:2009.11462},
  year    = {2020}
}

@article{birhane2021multimodal,
  author  = {Birhane, Abeba and Prabhu, Vinay Uday and Kahembwe, Emmanuel},
  title   = {Multimodal datasets: misogyny, pornography, and malignant stereotypes},
  journal = {arXiv preprint arXiv:2110.01963},
  year    = {2021}
}

@inproceedings{zhou2022feature,
  title={Do feature attribution methods correctly attribute features?},
  author={Zhou, Yilun and Booth, Serena and Ribeiro, Marco Tulio and Shah, Julie},
  booktitle={Proceedings of the AAAI conference on artificial intelligence},
  volume={36},
  number={9},
  pages={9623--9633},
  year={2022}
}

@inproceedings{yuksekgonul2023posthoc,
  title  = {Post-hoc Concept Bottleneck Models},
  author = {Mert Yuksekgonul and Maggie Wang and James Zou},
  booktitle = {The Eleventh International Conference on Learning Representations},
  year   = {2023},
  url    = {https://openreview.net/forum?id=nA5AZ8CEyow}
}

@InProceedings{pmlr-v280-wang25b,
  title     = {Concept Bottleneck Model with Zero Performance Loss},
  author    = {Wang, Zhenzhen and Popel, Aleksander and Sulam, Jeremias},
  booktitle = {Conference on Parsimony and Learning},
  pages     = {433--461},
  year      = {2025},
  editor    = {Chen, Beidi and Liu, Shijia and Pilanci, Mert and Su, Weijie and Sulam, Jeremias and Wang, Yuxiang and Zhu, Zhihui},
  volume    = {280},
  series    = {Proceedings of Machine Learning Research},
  month     = {24--27 Mar},
  publisher = {PMLR},
  url       = {https://proceedings.mlr.press/v280/wang25b.html}
}

@inproceedings{NEURIPS2024_8c1df815,
  author    = {Teneggi, Jacopo and Sulam, Jeremias},
  booktitle = {Advances in Neural Information Processing Systems},
  doi       = {10.52202/079017-2435},
  editor    = {A. Globerson and L. Mackey and D. Belgrave and A. Fan and U. Paquet and J. Tomczak and C. Zhang},
  pages     = {76450--76499},
  publisher = {Curran Associates, Inc.},
  title     = {Testing Semantic Importance via Betting},
  volume    = {37},
  year      = {2024}
}

@article{chattopadhyay2023variational,
  title   = {Variational Information Pursuit for Interpretable Predictions},
  author  = {Chattopadhyay, Aditya and Chan, Kwan Ho Ryan and Haeffele, Benjamin D and Geman, Donald and Vidal, Ren{\'e}},
  journal = {arXiv preprint arXiv:2302.02876},
  year    = {2023}
}

@article{cunningham2023sparse,
  title={Sparse autoencoders find highly interpretable features in language models},
  author={Cunningham, Hoagy and Ewart, Aidan and Riggs, Logan and Huben, Robert and Sharkey, Lee},
  journal={arXiv preprint arXiv:2309.08600},
  year={2023}
}
\bibliographystyle{icml2026}

%%%%%%%%%%%%%%%%%%%%%%%%%%%%%%%%%%%%%%%%%%%%%%%%%%%%%%%%%%%%%%%%%%%%%%%%%%%%%%%
%%%%%%%%%%%%%%%%%%%%%%%%%%%%%%%%%%%%%%%%%%%%%%%%%%%%%%%%%%%%%%%%%%%%%%%%%%%%%%%
% APPENDIX
%%%%%%%%%%%%%%%%%%%%%%%%%%%%%%%%%%%%%%%%%%%%%%%%%%%%%%%%%%%%%%%%%%%%%%%%%%%%%%%
%%%%%%%%%%%%%%%%%%%%%%%%%%%%%%%%%%%%%%%%%%%%%%%%%%%%%%%%%%%%%%%%%%%%%%%%%%%%%%%
\newpage
\appendix
\onecolumn

\appendix
\onecolumn
\begin{center}
\textbf{\large Supplementary Materials}
\end{center}
\setcounter{section}{0}
\renewcommand{\thesection}{\Alph{section}}

\section{
Theoretical Guarantees and Relation to Merlin--Arthur Classifiers}
\label{app:theory}

Neural Concept Verifier (NCV) builds on the Merlin--Arthur classifier (MAC) framework of
\cite{pmlr-v238-waldchen24a}, which provides information-theoretic interpretability guarantees in a
binary classification setting. In this appendix, we briefly recall these guarantees and explain how
they apply in our concept-based setting. We do \emph{not} prove new theorems here; rather, we
instantiate existing results and make explicit which assumptions are required and where our claims
remain empirical.

\subsection{Merlin--Arthur guarantees in the binary case}
\label{app:ma-binary}

We briefly recall the guarantees for Merlin--Arthur classifiers in the original
binary setting of \cite{pmlr-v238-waldchen24a}, focusing on the quantities that appear
in our NCV discussion: average precision, mutual information, completeness,
soundness, asymmetric feature correlation, and relative success rate.

\paragraph{Setup.}
Throughout this subsection, the notation is local to the binary Merlin--Arthur
setting of \cite{pmlr-v238-waldchen24a} and should not be conflated with the
concept-space notation of the main text; in particular, $\mathcal{D}$ here denotes
a distribution over data points (with a deterministic label map), rather than the
joint distribution over input--label pairs used in the main paper.
We consider a two-class data space $\mathfrak{D} = (\Omega, \mathcal{D}, \ell)$,
where $\Omega$ is the set of data points, $X \sim \mathcal{D}$, and
$\ell : \Omega \to \{-1, 1\}$ is a deterministic label map with $L = \ell(X)$.
The class-conditional distributions are $\mathcal{D}_s := \mathcal{D}(\,\cdot \mid L = s)$
for $s \in \{-1, 1\}$.
Following \cite{pmlr-v238-waldchen24a}, each data point $x \in \Omega$ is identified
with the set of its indexed entries $\{(1, x_1), \dots, (d, x_d)\}$, and a
\emph{feature} is a partial object $z \in \Omega_p$ (\eg, a subset of pixels);
we write $z \subseteq x$ when $z$ is contained in this set, \ie\ when $x$ exhibits
the feature $z$.
A feature selector maps a data point to one of its features, $M(x) \subseteq x$;
we write $M$ for Merlin and $\widehat{M}$ for Morgana.
Arthur is a classifier $A$ predicting in $\{-1, \bot, 1\}$, where $\bot$ denotes
abstention (the ``Don't know'' option).

\paragraph{Average precision and mutual information.}
Given a feature $z$ and a reference point $x'$ that exhibits it ($z \subseteq x'$),
the \emph{precision} of $z$ is the probability that an independently drawn point
$x \sim \mathcal{D}$ which also exhibits $z$ shares the label of $x'$:
\[
  \Pr(z; x')
  := \mathbb{P}_{x\sim\mathcal{D}}\big[\ell(x) = \ell(x') \mid z \subseteq x \big]
  \,.
\]
For a feature selector $M$, the \emph{average precision} is the expected precision
of the features it selects:
\begin{equation}
  \Pr_{\mathcal{D}}(M)
  :=
  \mathbb{E}_{x'\sim\mathcal{D}}
  \Big[
    \mathbb{P}_{x\sim\mathcal{D}}\big[\ell(x) = \ell(x') \mid M(x') \subseteq x\big]
  \Big].
  \label{eq:avg-precision}
\end{equation}
This quantity bounds the average conditional entropy of the class given Merlin's
features, and hence the mutual information. Writing $H_b$ for the binary entropy
and $H(\cdot)$, $I(\cdot\,;\cdot)$ for entropy and mutual information,
\cite{pmlr-v238-waldchen24a} show
\begin{equation}
  \mathbb{E}_{x'\sim\mathcal{D}}
  \big[
    I_{x\sim\mathcal{D}}\big(\ell(x);\, M(x') \subseteq x\big)
  \big]
  \;\ge\;
  H_{x\sim\mathcal{D}}\big(\ell(x)\big)
  - H_b\big(\Pr_{\mathcal{D}}(M)\big),
  \label{eq:mi-bound-precision}
\end{equation}
where, for fixed $x'$, $M(x') \subseteq x$ denotes the binary containment variable.
As $\Pr_{\mathcal{D}}(M) \to 1$, the term $H_b(\Pr_{\mathcal{D}}(M)) \to 0$ and
Merlin's features carry almost all available label information in this binary
setting.

\paragraph{Idealised min--max guarantee (optimal players).}
In the idealised setting, Arthur and Morgana are assumed to play optimally
against a fixed Merlin. Define the error set
\[
  E_{M, \widehat{M}, A}
  :=
  \Big\{
    x \in \Omega
    \,\big|\,
    A(M(x)) \neq \ell(x)
    \;\lor\;
    A(\widehat{M}(x)) = -\ell(x)
  \Big\},
\]
\ie, the points where Merlin fails to convince Arthur of the correct label or
Morgana forces an incorrect (non-abstaining) prediction. The min--max error of
Merlin is
\[
  \varepsilon_M
  :=
  \min_{A}
  \max_{\widehat{M}}
  \mathbb{P}_{x\sim\mathcal{D}}\big[x \in E_{M, \widehat{M}, A}\big].
\]
Theorem 2.7 of \cite{pmlr-v238-waldchen24a} states that if $\varepsilon_M$ is small,
there exists a subset $\Omega' \subseteq \Omega$ of mass at least $1-\varepsilon_M$
such that, on the induced data space $\mathfrak{D}' = (\Omega', \mathcal{D}', \ell)$,
Merlin achieves perfect precision, and hence zero conditional class entropy:
\[
  \Pr_{\mathcal{D}'}(M) = 1
  \quad\text{and thus}\quad
  H_{x',x\sim\mathcal{D}'}\big(\ell(x') \mid M(x')\subseteq x\big) = 0.
\]
In words: if an optimal Arthur--Morgana pair can almost never disagree with Merlin,
then on almost all of the data space Merlin's features determine the label uniquely.

\paragraph{Realistic players: completeness, soundness, AFC, and relative strength.}
For high-dimensional data, exhaustive search for an optimal Morgana is not
feasible. The analysis is therefore relaxed to \emph{realistic} (\eg, neural)
players and expressed in terms of:
\begin{itemize}
  \item \textbf{Completeness}
    \[
      \min_{s\in\{-1,1\}}
      \mathbb{P}_{x\sim\mathcal{D}_s}\big[A(M(x)) = \ell(x)\big]
      \;\ge\; 1 - \varepsilon_c,
    \]
    \ie, Merlin's features let Arthur classify correctly with high probability
    in each class.
  \item \textbf{Soundness}
    \[
      \max_{s\in\{-1,1\}}
      \mathbb{P}_{x\sim\mathcal{D}_s}\big[A(\widehat{M}(x)) = -\ell(x)\big]
      \;\le\; \varepsilon_s,
    \]
    \ie, Morgana almost never forces a wrong prediction; on her subsets, Arthur
    either stays correct or abstains (predicts $\bot$).
  \item \textbf{Class imbalance $B$}, bounding how skewed the class prior can be,
    \[
      B := \max_{s\in\{-1,1\}}
      \frac{\mathbb{P}_{x\sim\mathcal{D}}[\ell(x) = s]}{\mathbb{P}_{x\sim\mathcal{D}}[\ell(x) = -s]}.
    \]
  \item \textbf{Asymmetric Feature Correlation (AFC) $\kappa$}, which measures how
    strongly a set of features can be \emph{concentrated} in a few points of one
    class while \emph{spread out} across many points of the other. A large $\kappa$
    lets Merlin use globally uninformative features (appearing equally often in
    both classes) while still attaining high completeness and soundness. We use
    $\kappa$ only through the assumption $\kappa = O(1)$ and refer to Definition~2.8
    of \cite{pmlr-v238-waldchen24a} for its formal definition.
  \item \textbf{Relative success rate $\alpha$} of Morgana, comparing how often she
    finds a convincing feature in the \emph{wrong} class to how often Merlin does
    so in the \emph{correct} class, restricted to points where such a feature
    exists:
    \[
      \alpha :=
      \min_{s\in\{-1,1\}}
      \frac{
        \mathbb{P}_{x\sim\mathcal{D}_{-s}}\big[A(\widehat{M}(x)) = s \mid x\in F_s^\ast\big]
      }{
        \mathbb{P}_{x\sim\mathcal{D}_s}\big[A(M(x)) = s \mid x\in F_s^\ast\big]
      },
    \]
    where $F_s^\ast := \{x \in \Omega \mid \exists\, z \subseteq x : z \in M(\mathcal{D}_s),\, A(z) = s\}$
    is the set of points containing a feature that Merlin selects on some class-$s$
    point and that Arthur accepts as class $s$. Intuitively, $\alpha$ is large when
    Morgana's search is at least as powerful as Merlin's.
\end{itemize}

Under these conditions, \cite{pmlr-v238-waldchen24a} prove that completeness,
soundness, AFC, class imbalance, and relative success rate jointly lower-bound the
average precision:
\begin{equation}
  \Pr_{\mathcal{D}}(M)
  \;\ge\;
  1 - \varepsilon_c
  -
  \frac{\kappa \alpha^{-1} \varepsilon_s}{
    1 - \varepsilon_c + \kappa \alpha^{-1} B^{-1} \varepsilon_s
  }.
  \label{eq:precision-bound}
\end{equation}
Furthermore, the AFC is bounded by the maximum number of features per data point:
if each point contains at most $F_{\max}$ features of the type considered (\eg,
patches of a fixed size), then $\kappa \le F_{\max}$, which gives a concrete regime
in which the assumption $\kappa = O(1)$ holds.

Combining the precision bound in \autoref{eq:precision-bound} with the
mutual-information bound in \autoref{eq:mi-bound-precision} yields
\[
  \mathbb{E}_{x'\sim\mathcal{D}}
  \big[
    I_{x\sim\mathcal{D}}\big(\ell(x);\, M(x')\subseteq x\big)
  \big]
  \;\ge\;
  H_{x\sim\mathcal{D}}\big(\ell(x)\big)
  - H_b\!\left(
      1 - \varepsilon_c
      - \frac{\kappa\alpha^{-1}\varepsilon_s}{
          1 - \varepsilon_c + \kappa\alpha^{-1}B^{-1}\varepsilon_s}
    \right),
\]
with $M(x')\subseteq x$ understood as above. For \emph{balanced} datasets
($B \approx 1$), bounded AFC ($\kappa = O(1)$), and a reasonably strong Morgana
($\alpha = O(1)$), high completeness ($\varepsilon_c \ll 1$) and soundness
($\varepsilon_s \ll 1$) therefore imply that Merlin's features carry almost all
label information in this binary setting.

\subsection{Instantiation for concept-based models}
\label{app:ncv-instantiation}
NCV applies the same prover--verifier game as Merlin--Arthur classifiers, but in a
\emph{concept space} rather than pixel space, and in a \emph{multi-class} rather than
binary setup. We return here to the notation of the
main text (\autoref{sec:setup}): for an input $x \in \mathbb{R}^D$, a concept
extractor $g$ produces an encoding $\mathbf{c} = g(x) \in \mathbb{R}^C$.
In analogy to the index--value features of \autoref{app:ma-binary}, a feature here is
a concept index $j$ together with its value $g(x)_j$; Merlin selects up to $m$ such
(concept, value) pairs via the mask $M(g(x))$. Merlin and Morgana
are implemented as provers $M, \widehat{M}: \mathbb{R}^C \to \{0,1\}^C$ that output
sparse binary masks $M(\mathbf{c}), \widehat{M}(\mathbf{c})$ with at most $m$ active
entries. Arthur then predicts using the masked encodings
\[
  S = M(\mathbf{c}) \odot \mathbf{c}
  \quad\text{and}\quad
  \widehat{S} = \widehat{M}(\mathbf{c}) \odot \mathbf{c}.
\]

We treat the selected partial encoding $M(g(x')) \odot g(x')$ as Merlin's feature, and
define the containment relation $M(x') \subseteq x$ (writing $M(x') := M(g(x'))$ as in
the main text) to hold when $g(x)_i = g(x')_i$ at every index $i$ with $M(g(x'))_i = 1$,
\ie\ when $x$ and $x'$ agree on every concept Merlin selects from $x'$. Under this
instantiation, the binary results of \autoref{app:ma-binary} apply to each one-vs-rest
subproblem, with concept-space parameters $(\kappa_k, \alpha_k, B_k)$, subject to the
evaluability caveat discussed below.

To connect this to the binary setting of \autoref{app:ma-binary}, we apply a
one-vs-rest reduction: for a fixed class $k \in \{1,\dots,K\}$ we form the binary task
with label $Y_k(y) = +1$ if $y = k$ and $-1$ otherwise (class $k$ vs.\ all others). The
selector $M$, the feature $M(g(x))\odot g(x)$, and Arthur (now predicting in
$\{-1,\bot,1\}$) play the roles of the corresponding objects in \autoref{app:ma-binary},
with the containment relation as defined above.

Let $\Pr_{\mathcal{D}}^{(k)}(M)$ denote the \emph{average precision} of Merlin's concept
features on this one-vs-rest task. As in \autoref{sec:theory} we work over input--label pairs
$(x,y),(x',y') \sim \mathcal{D}$, and set, analogously to \autoref{eq:avg-precision},
\[
  \Pr_{\mathcal{D}}^{(k)}(M)
  :=
  \mathbb{E}_{(x',y')}
  \Big[
    \mathbb{P}_{(x,y)}\big[Y_k(y) = Y_k(y') \mid M(x') \subseteq x\big]
  \Big].
\]
Applying the mutual-information bound of \autoref{eq:mi-bound-precision} to the binary
label $Y_k(y)$ and selector $M$ yields
\begin{equation}
  \mathbb{E}_{(x',y')\sim\mathcal{D}}
  \big[
    I_{(x,y)\sim\mathcal{D}}\big(Y_k(y);\, M(x') \subseteq x\big)
  \big]
  \;\ge\;
  H_{y\sim\mathcal{D}}\big(Y_k(y)\big) - H_b\big(\Pr_{\mathcal{D}}^{(k)}(M)\big),
  \label{eq:mi-bound-concepts}
\end{equation}
where, for fixed $x'$, $M(x') \subseteq x$ denotes the binary containment variable.
Thus, if $\Pr_{\mathcal{D}}^{(k)}(M)$ is close to $1$, Merlin's sparse concept subsets
for class $k$ carry almost all information about $Y_k(y)$.

The analysis of \autoref{app:ma-binary} further relates $\Pr_{\mathcal{D}}^{(k)}(M)$ to
\emph{observable} completeness and soundness on this binary subproblem, under three
additional assumptions:
\begin{itemize}
  \item a bounded \emph{Asymmetric Feature Correlation} (AFC) $\kappa_k$ for the
    one-vs-rest task of class $k$, measured in concept space. When a single constant is
    convenient we use the uniform bound $\kappa := \max_k \kappa_k$. If each input has
    at most $m_{\max}$ active concepts (\eg, under a fixed sparsity level or top-$m$
    selection), then $\kappa_k \le m_{\max}$, giving a concrete regime in which the
    assumption $\kappa_k = O(1)$ holds.
  \item a bounded \emph{class imbalance} $B_k$ for the one-vs-rest task;
  \item a \emph{relative success rate} $\alpha_k > 0$ of Morgana, \ie\ her search over
    concept subsets is roughly as powerful as Merlin's. In the ideal full-search
    setting $\alpha_k = 1$; in NCV we use symmetric neural architectures for Merlin and
    Morgana as heuristic evidence that $\alpha_k$ is close to $1$, but we do not
    estimate it explicitly.
\end{itemize}

Under these conditions, the precision bound \autoref{eq:precision-bound} applies
class-wise: for each $k$ one obtains
\[
  \Pr_{\mathcal{D}}^{(k)}(M)
  \;\ge\;
  1 - \varepsilon_c^{(k)}
  -
  \frac{\kappa_k \alpha_k^{-1} \varepsilon_s^{(k)}}{
    1 - \varepsilon_c^{(k)} + \kappa_k \alpha_k^{-1} B_k^{-1} \varepsilon_s^{(k)}
  },
\]
where $\varepsilon_c^{(k)}$ and $\varepsilon_s^{(k)}$ are the completeness and soundness
errors of NCV on the one-vs-rest problem for class $k$. Combining this with
\autoref{eq:mi-bound-concepts} yields a lower bound on the mutual information between the
class-$k$ label and Merlin's concept subsets.

In NCV we view these results as an \emph{idealised} description of the concept-level
prover--verifier game:
\begin{itemize}
  \item On each one-vs-rest task, and under the AFC and relative-strength assumptions
    above, high completeness and soundness imply that Merlin's sparse concept subsets
    are highly informative about the label in the sense of \autoref{eq:mi-bound-concepts}.
  \item The precision $\Pr_{\mathcal{D}}^{(k)}(M)$ is well-defined for any concept
    extractor, but estimating it empirically requires the selected concepts to recur
    across inputs. This holds for discrete or low-dimensional encodings (\eg, our NCB
    instantiation), but for dense continuous encodings such as CLIP exact recurrence is
    vanishingly rare, and the precision is then meaningful only as a population
    quantity. The bound is therefore most directly evaluable in the discrete,
    moderate-$C$ regime.
  \item In our high-dimensional, multi-class experiments we do not estimate the
    class-wise parameters $(\varepsilon_c^{(k)}, \varepsilon_s^{(k)}, \alpha_k, B_k, \kappa_k)$
    explicitly. We therefore interpret the Merlin--Arthur theory as a \emph{theoretical
    lens} for NCV, rather than a quantitative certification for each dataset;
    empirically reported completeness and soundness should be read in this light
    (\cf the discussion in \autoref{sec:theory}).
\end{itemize}
\subsection{Notion of faithfulness}
\label{app:faithfulness}

\paragraph{Definition (Faithfulness of \method).}
For a class $k \in \{1, \dots, K\}$, we say that \method produces \emph{faithful}
explanations in the information-theoretic sense inherited from Merlin--Arthur (and
adapted to concept space) when the following three conditions hold:
\begin{enumerate}
  \item \textbf{Completeness.} For an input $x$ with true class $k$, Merlin produces a
    sparse concept subset $S(x) = M(\mathbf{c}) \odot \mathbf{c}$ (with $\mathbf{c} = g(x)$)
    such that, with high probability over $x \sim \mathcal{D}_k$, Arthur can correctly
    predict $k$ from $S(x)$ alone; \ie, the completeness error $\varepsilon_c^{(k)}$ for
    the corresponding one-vs-rest task is small.
  \item \textbf{Soundness.} Morgana cannot find alternative concept subsets that force
    Arthur into a wrong prediction; at worst, Arthur abstains via the rejection class
    $\bot$. Equivalently, the soundness error $\varepsilon_s^{(k)}$ in the sense of
    \autoref{app:ma-binary} is small.
  \item \textbf{Precision-bound assumptions.} The concept-space AFC parameter $\kappa_k$,
    the class imbalance $B_k$, and Morgana's relative success rate $\alpha_k$ are
    well-behaved (bounded, with $\alpha_k$ not too small), so that the precision bound
    of \autoref{app:ma-binary} links the errors $\varepsilon_c^{(k)}, \varepsilon_s^{(k)}$
    to the class-wise average precision $\Pr_{\mathcal{D}}^{(k)}(M)$.
\end{enumerate}
When all three hold, the precision bound guarantees that $\Pr_{\mathcal{D}}^{(k)}(M)$ is
close to $1$, and the mutual-information bound \autoref{eq:mi-bound-concepts} then
implies that Merlin's sparse concept subsets for class $k$ carry near-maximal
information about $Y_k(y)$.

In our experiments we report completeness and soundness empirically, but we do not
attempt to estimate the resulting lower bounds on $\Pr_{\mathcal{D}}^{(k)}(M)$, for the
reasons detailed in \autoref{app:limitations}. This is explicitly \emph{not} a causal
guarantee: NCV does not show that the concepts are causally sufficient for the task. It
ensures only that, relative to the given concept representation and under the
assumptions above, the sparse subsets Merlin selects are informative and robust in the
sense made precise by the bounds above.

\subsection{Use of the rejection class and training objective}
\label{app:rejection-loss}
Finally, we clarify how the rejection class is used during training, since it is crucial
for enforcing soundness in the sense of \autoref{app:ma-binary}. Arthur outputs logits
$z = A(S) \in \mathbb{R}^{K+1}$ over $K$ classes plus a rejection class $\bot$, and we
write $p = \mathrm{softmax}(z)$ for the predictive probabilities, with $p_y$ and $p_\bot$
the probabilities of the true class $y$ and the rejection class. Given a concept encoding
$\mathbf{c} = g(x)$ and the Merlin/Morgana subsets $S = M(\mathbf{c}) \odot \mathbf{c}$ and
$\widehat{S} = \widehat{M}(\mathbf{c}) \odot \mathbf{c}$, we use the following losses.

\paragraph{Merlin loss.}
\[
  L_M = -\log p_y,
\]
i.e., standard cross-entropy with respect to the true class on Merlin's subset.

\paragraph{Morgana loss (soundness).}
Morgana's loss operates directly on the logits for the true class and the rejection
class. Let $\widehat{z} = A(\widehat{S}) \in \mathbb{R}^{K+1}$ be Arthur's logits on
Morgana's subset, with components $\widehat{z}_y$ and $\widehat{z}_\bot$. We first define
a modified target $\tilde{y}$ that switches to the rejection class whenever Arthur already
prefers $\bot$ over $y$ on $\widehat{S}$:
\[
  \tilde{y}
  =
  \begin{cases}
    y,    & \text{if } \widehat{z}_y \ge \widehat{z}_\bot,\\
    \bot, & \text{if } \widehat{z}_\bot > \widehat{z}_y,
  \end{cases}
\]
and set
\[
  L_{\widehat{M}}
  =
  \mathrm{CE}\big(\widehat{z}, \tilde{y}\big)
  \;+\;
  \frac{1}{N}\sum_{i=1}^{N}
    \Big(-\log\big(1 + \exp(-\lvert \widehat{z}_y^{(i)} - \widehat{z}_\bot^{(i)}\rvert)\big)\Big),
\]
where $N$ is the batch size. Arthur minimises this loss while Morgana maximises it. The
cross-entropy term targets the correct-or-abstain label $\tilde{y}$, and the second term
is a softplus penalty on the gap $\lvert \widehat{z}_y - \widehat{z}_\bot\rvert$. Through
the $\bot$ branch of $\tilde{y}$, abstention counts as sound, consistent with
\autoref{app:ma-binary}.

\paragraph{Arthur's objective and optimisation.}
Arthur's overall loss is the convex combination
\[
  L_A = (1-\gamma)\, L_M + \gamma\, L_{\widehat{M}},
  \qquad \gamma \in [0, 0.5],
\]
trading off completeness and soundness; all losses use mean reduction over the batch.
Although the concept selections are ultimately discrete, both provers are trained through
continuous masks. Each prover outputs real-valued scores
$m_{\mathrm{cont}}, \widehat{m}_{\mathrm{cont}} \in \mathbb{R}^C$, used as soft masks to
form
\[
  S_{\mathrm{soft}} = m_{\mathrm{cont}} \odot \mathbf{c},
  \qquad
  \widehat{S}_{\mathrm{soft}} = \widehat{m}_{\mathrm{cont}} \odot \mathbf{c}.
\]
When updating the provers (Merlin by gradient descent on $L_M$, Morgana by gradient ascent
on $L_{\widehat{M}}$), Arthur is frozen and fed $S_{\mathrm{soft}}, \widehat{S}_{\mathrm{soft}}$,
so that gradients flow back into the continuous mask parameters. After the prover updates we
discretise the masks with a top-$m$ operator, setting the $m$ entries of largest magnitude
in $m_{\mathrm{cont}}$ (respectively $\widehat{m}_{\mathrm{cont}}$) to $1$ and the rest to
$0$, yielding hard masks $M(\mathbf{c}), \widehat{M}(\mathbf{c}) \in \{0,1\}^C$. Arthur is
then updated by gradient descent on $L_A$ using the hard-masked encodings $S$ and
$\widehat{S}$. This alternating scheme, namely continuous masks for gradient flow in the
provers and hard top-$m$ masks for Arthur's update, implements a practical min-max
training procedure for NCV. The explicit rejection class $\bot$ ensures that soundness
measures robustness to \emph{adversarial} concept selections, in that Arthur is never
driven to be confidently wrong, aligning with the Merlin--Arthur framework instantiated in
concept space.

\subsection{Limitations of the theoretical guarantees}
\label{app:limitations}

For completeness, we also spell out the main limitations of the information-theoretic
guarantees when applied to NCV's high-dimensional, multi-class, concept-based
setting.

First, the original theory is formulated for binary classification. Our
instantiation in \autoref{app:ncv-instantiation} uses a one-vs-rest reduction
to obtain class-wise guarantees on the mutual information
$I(Y_k(y);\, M(x') \subseteq x)$ for each $k$, but it does not provide a direct
statement about the joint $K$-class decision or about the final
$\arg\max$ prediction over all classes.

\paragraph{High-dimensional sparsity and feature reuse.}
A central limitation arises from the extreme sparsity of Merlin's and Morgana's
selections in our high-dimensional concept spaces. In the original
Merlin--Arthur setting, features are typically small, localized structures
(\eg, image patches) and the experiments are conducted on relatively
low-dimensional, small-scale datasets. In this regime, many inputs share the
same features, so the event $\{M(x') \subseteq x\}$ has non-negligible
probability and the average precision $\Pr_{\mathcal{D}}(M)$ in~\autoref{eq:avg-precision} can be meaningfully interpreted and
estimated from finite samples. In NCV, Merlin and Morgana select small
subsets of a large concept vocabulary (\eg, $m \ll C$ with $C$ in the thousands),
and in practice the exact subsets $M(x')$ and $\widehat{M}(x')$ are often highly
specific to each input. As a result, the precise event $\{M(x') \subseteq x\}$
may have very low probability under $\mathcal{D}$, and empirical estimates of
$\Pr_{\mathcal{D}}^{(k)}(M)$ become unstable in finite samples. For this reason,
we do not attempt to estimate average precision or the resulting mutual-information
lower bounds numerically in our experiments. Instead, we use completeness and
soundness as observable proxies and treat the precision/MI guarantees as an
idealised, distribution-level description of the behaviour that NCV is designed
to promote, rather than as directly calibrated finite-sample certificates.

\paragraph{Class imbalance in one-vs-rest reductions.}
In large-$K$ settings such as ImageNet-1k, the one-vs-rest subproblems for each
class $k$ are highly imbalanced (the positive prior is approximately $1/K$).
In the precision bound of~\autoref{app:ncv-instantiation}, this is reflected
in the imbalance parameter $B_k$, whose large value makes the resulting lower
bound on $\Pr_{\mathcal{D}}^{(k)}(M)$ more conservative: small soundness errors
$\varepsilon_s^{(k)}$ are penalised more strongly relative to completeness
$\varepsilon_c^{(k)}$. This provides an additional reason why we do not attempt
to compute numerical mutual-information lower bounds in our experiments, and
instead interpret the Merlin--Arthur theory qualitatively as a guiding framework.

\paragraph{Relative success and adversary class.}
Finally, the robustness interpretation of soundness in our setting is tied to the particular adversarial prover class we train in practice (a neural network $\widehat{M}$ with a specific architecture and loss). In the Merlin--Arthur
framework, this dependence is captured abstractly via the relative success rate
$\alpha$, which measures how powerful Morgana is compared to Merlin on those
points where Merlin can provide convincing evidence. In NCV we use symmetric
neural architectures for $M$ and $\widehat{M}$ as heuristic evidence that
$\alpha_k$ is not tiny, but we do not attempt to verify or optimise $\alpha_k$
beyond this design choice. As a result, our empirical soundness estimates should
be interpreted as robustness against this trained adversary class, rather than
against an arbitrary worst-case adversary over all admissible subsets. This is
fully in line with the practical use of the Merlin--Arthur framework in
\cite{pmlr-v238-waldchen24a}, where the general theory is instantiated and
evaluated with specific neural implementations of Merlin, Morgana, and Arthur.

Overall, we see the Merlin--Arthur framework as providing a rigorous
\emph{idealised model} for the kind of behaviour NCV is designed to encourage:
high completeness, robustness to adversarial concept subsets, and information-rich
sparse explanations in concept space. In our experiments, we use completeness
and soundness as observable proxies for these properties, but we do not claim
fully certified guarantees beyond the stated assumptions.

\section{Why Linear Classifiers Fall Short in CBMs}\label{app:linear}
While linear classifiers are generally considered to be interpretable, these models are not suited to solve arbitrarily complex problems. A linear classifier is only able to capture linear relationships between inputs and output features and cannot model complex, non-linear relationships. In the context of CBMs, this problem is usually tackled by utilizing a linear classifier to predict the output based on the detected concepts. The concepts themselves can be detected using non-linear models, and only the classification based on these concepts is done with a linear model. However, this is not always sufficient, as there are also simple examples where non-linear relationships between concepts and the output exist, for example thresholds detection (three out of five symptoms need to be present to indicate an illness) or multiplicative effects (crop yield is the result of a multiplicative relationship between rain and fertility).

To illustrate the problem in a simple experimental setup, let us assume we have a dataset of simple shapes and every image contains between one and four of these shapes. The shapes are either a square or a circle and either orange or blue. We consider two simple classification scenarios for this dataset.

\begin{itemize}
    \item \textbf{XOR:} This setting classification follows the traditional XOR problem: We want to classify images that contain either an orange square or a blue circle as class one and all other images as class two.
    \item \textbf{Counting:} This setting includes object counting and illustrates that even for classification based on a single attribute, a linear layer can be insufficient. Here, we want to classify all images with exactly one blue shape as class one, and all other images as class two.
\end{itemize}
We evaluate a linear layer and a simple MLP on this toy dataset. To further simplify things, we assume that our concept encoder is able to perfectly detect the concepts in the image, thus providing for each element the information whether there is an object and if so, its shape and color. We randomly generate 5000 samples of the dataset and train the models on a train split of 80\% and evaluate on the remaining 20\%. The MLP has one hidden layer of size 16 and uses ReLU activation functions.

\begin{wraptable}{r}{0.55\linewidth}
    \vspace{-10pt}

    \caption{A linear prediction layer cannot solve \textit{XOR} or \textit{counting}. Even with the assumption of a perfect concept encoder, the linear layer fails.}
    \vspace{0.5em}
    \centering
    \begin{tabular}{lcc}
        \toprule
        Model & XOR (Acc) & Counting (Acc)\\
        \midrule
        Linear Layer & $0.766 \pm 0.011$ & $0.677 \pm 0.006$\\
        MLP & $0.953 \pm 0.053$ & $0.982 \pm 0.015$\\
        \bottomrule
    \end{tabular}
    \label{tab:illustrating_example}
    \vspace{-5pt}
\end{wraptable}

The results of this evaluation are shown in \autoref{tab:illustrating_example}. In both scenarios, the linear classification layer is not able to solve the task, despite the deceptively simple relationship between concepts and output classes. On the other hand, the MLP achieves close to perfect accuracy on both settings.

So far, we have argued that not every task can be solved with a CBM and a linear classification layer. However, this is not entirely accurate. In principle, any task can be solved linearly---provided that we define the right \textit{linear-sufficient} concepts. For instance, in the XOR setting, detecting the concepts ``orange square and no blue circle'' and ``blue circle and no orange square'' would allow a linear classifier to solve the task. Similarly, in the counting task, introducing a concept such as ``exactly one blue object'' would make linear classification trivial.  

That said, the assumption that such sufficient concepts are always available is not realistic. First, designing or discovering these concepts often makes concept detection considerably more difficult. Second, as concepts become increasingly specific and compositional, they tend to lose interpretability. Finally, requiring tailored concepts for every individual task does not scale. Returning to our example, the concept ``exactly one blue object'' might help with task two but is essentially useless for task one.  

Taken together, this illustrates why relying solely on linear classifiers in CBMs is often impractical. To address such cases, non-linear classifiers should also be considered.

\section{Experimental Details: Baselines}\label{app:baselines}

In this section, we provide training details for the considered baselines: \textit{ResNet-18}, \textit{ResNet-50}, \textit{Pixel-MAC}, \textit{CBM} (linear and nonlinear variants), and \textit{DCR}.

\subsection{ResNet-18 and ResNet-50}
We initialize the framework with a pretrained ResNet-18 model and employ the Adam optimizer across all experiments. On the CLEVR datasets, the model is trained with a batch size of 128 for 30 epochs using a learning rate of $10^{-4}$ and weight decay of $10^{-4}$, repeated across 20 random seeds with early stopping based on validation loss.  

On CIFAR-100, we use a ResNet-50 trained for 100 epochs with a learning rate of $10^{-4}$, weight decay of $10^{-5}$, and a batch size of 128, averaged over 10 random seeds with early stopping. The ResNet-50 baseline on ImageNet is evaluated directly using pretrained PyTorch \citep{paszke2019pytorch} weights without further finetuning. For COCOLogic, a ResNet-50 is trained for 300 epochs with a batch size of 256, learning rate of $10^{-4}$, and weight decay of $10^{-2}$, again averaged over 10 random seeds with early stopping.  

In the Pixel-MAC setup, a separate ResNet-18 is trained under the same configuration as above but with a reduced learning rate of $10^{-5}$, while keeping the batch size, weight decay, and early stopping criterion unchanged. All Pixel-MAC results are obtained from these ResNet-18 checkpoints.

\subsection{Pixel-MAC}
In this setup, we apply Merlin-Arthur training on pixel space by utilizing the pretrained ResNet-18 models as classifiers and U-Net architectures for both Merlin and Morgana. Throughout all experiments, we employ the Adam optimizer for both classifier and feature selector optimization, with $\gamma=0.5$ to ensure high soundness. 

For the CLEVR datasets, we train with a batch size of 128 for 40 epochs, using a learning rate of $10^{-5}$ and weight decay of $10^{-6}$ for the classifier optimization. The U-Net architectures are trained with a learning rate of $10^{-4}$, weight decay of $10^{-5}$ and an L1 penalty coefficient of $0.1$. We set the mask size to 1500, meaning that the U-Nets select a subset of 1500 pixels per sample (out of $128\!\times\!128$ pixels).

For CIFAR-100, we reduce the batch size to 64 and train for 100 epochs. The classifier is optimized with a learning rate of $10^{-5}$ and weight decay of $10^{-4}$. Both Merlin and Morgana are trained using a learning rate of $10^{-3}$ and a reduced mask size of 32 pixels. We use an L1 penalty of $0.01$ and apply early stopping across 10 random seeds.

For ImageNet and COCOLogic, we further reduce the batch size to 32 due to memory constraints and train for 80 epochs using the same learning rates and hyperparameters as in the CIFAR-100 setting. The mask size is set to 1000 pixels per image. As with CIFAR-100, early stopping is applied across 10 random seeds, and training is initialized from the pretrained ResNet-18 backbone.

\subsection{CBM with Linear Classifier}

Next, we present the implementation details for the CBM baseline, where a linear classifier operates on concept features obtained from the concept extractor. 

For the CLEVR datasets, we train a linear classifier on concepts extracted by the Neural Concept Binder. The training process employs a batch size of 128, a learning rate of $10^{-3}$, and weight decay of $10^{-4}$. The model is trained for 60 epochs on CLEVR-Hans3 and 30 epochs on CLEVR-Hans7, using early stopping based on validation loss, repeated across 20 different random seeds.

For CIFAR-100 and ImageNet, we train a linear classifier on sparse SpLiCE encodings using a dictionary size of 10,000. In both cases, we use a batch size of 4096 and train for 250 epochs with early stopping, a learning rate of $10^{-3}$, and no weight decay. A hidden layer with 512 units is used, and an L1 penalty of 0.2 is applied within SpLiCE to encourage sparsity in the concept representations. All results are averaged over 10 random seeds.

\subsection{DCR Baseline Implementation}
\label{app:dcr-details}

For the additional nonlinear CBM comparison in \autoref{sec:experimental_eval}, we implemented Deep Concept Reasoning (DCR; \citealt{barbiero2023interpretable}) following the official \texttt{torch-explain} library. Our implementation uses the same experimental setup as NCV to ensure a fair comparison.

\paragraph{Architecture.} We employ a ResNet-18 backbone (pretrained weights disabled for consistency) followed by two DCR-specific modules: (1) a \texttt{ConceptEmbedding} layer that maps backbone features to concept predictions $\mathbf{c}_{\text{pred}} \in [0,1]^{C}$ and learned concept embeddings $\mathbf{c}_{\text{emb}} \in \mathbb{R}^{C \times d}$, and (2) a \texttt{ConceptReasoningLayer} that learns class-specific logic rules in Disjunctive Normal Form (DNF) over the concept embeddings to produce class predictions.

\paragraph{Concept Supervision.} We supervise the concept predictions using the same CLIP-Sim similarity vectors as NCV. Since these are cosine similarities in $[-1, 1]$, we apply a linear normalization $c_{\text{truth}} = (\text{sim} + 1)/2$ to map them to $[0, 1]$ for BCE loss compatibility.

\paragraph{Training.} Following the official DCR tutorial, we optimize a joint loss $\mathcal{L} = \mathcal{L}_{\text{concept}} + \lambda \cdot \mathcal{L}_{\text{task}}$, where both terms use binary cross-entropy (BCE). The task loss uses one-hot encoded labels, as the \texttt{ConceptReasoningLayer} outputs per-class probabilities (not logits). We train with Adam optimizer for 100 epochs.

\paragraph{Hyperparameter Sweep.} 
For each dataset, we performed a grid search over:
\begin{itemize}[nosep]
    \item Vocabulary size: $n \in \{1000, 3000, 10000\}$
    \item Concept embedding dimension: $d \in \{8, 16, 32\}$
    \item Learning rate: $\eta \in \{10^{-3}, 10^{-4}\}$
    \item Task loss weight: $\lambda \in \{0.5, 1.0\}$
\end{itemize}
This resulted in $3 \times 3 \times 2 \times 2 = 36$ configurations per dataset, totaling over 100 training runs. For COCOLogic-10, we report balanced accuracy due to class imbalance.

\paragraph{Results.} \autoref{tab:dcr_comparison} summarizes the best accuracies achieved by DCR compared to NCV at 10k vocabulary size. DCR successfully trains on CIFAR-100 and COCOLogic-10 but fails to complete training on ImageNet-1k within reasonable time ($>3$ days per epoch). In contrast, NCV's sparsity-inducing mechanism enables efficient scaling to all three datasets under the same 10k-concept space.

\begin{table}[t]
    \centering
    \caption{Comparison between DCR and NCV at 10k-concept vocabulary size on CLIP-Sim. NCV metrics correspond to completeness (accuracy). For COCOLogic-10, we report balanced accuracy due to class imbalance.}
    \label{tab:dcr_comparison}
    \begin{tabular}{lcc}
        \toprule
        Dataset & DCR & \textbf{NCV} (ours) \\
                & (acc.~\%) & (completeness, \%) \\
        \midrule
        CIFAR-100    & 51.76 & \textbf{83.32} \\
        ImageNet-1k  & n/a (timeout)   & \textbf{67.04} \\
        COCOLogic-10 (bal.) & 38.61 & \textbf{75.42} \\
        \bottomrule
    \end{tabular}
\end{table}

\subsection{Comparison with Variational Information Pursuit (V-IP)}\label{app:vip}

Variational Information Pursuit (V-IP)~\citep{chattopadhyay2023variational} is closely related to \method in its information-theoretic motivation: both methods aim to identify a small set of features that are maximally informative about the label. In this section we discuss the methodological differences and provide a representative numerical comparison on CIFAR-100.

\paragraph{Method.} V-IP starts from an empty history and iteratively selects the next query whose answer is expected to be most informative about the label, conditional on the history collected so far. The query strategy and a classifier are parameterised by neural networks and trained end-to-end to minimise the expected cross-entropy between the true and predicted label. At inference time, V-IP issues queries sequentially until a confident prediction can be made, yielding an interpretable decision chain of variable length per input. In its original form, queries are realised as image patches (\eg, $8\times 8$ patches with stride $4$ on CIFAR, yielding $49$ candidate queries). Subsequent extensions also consider semantic queries from task-specific vocabularies.

\method differs in two main ways. First, \method is \emph{single-shot}: Merlin selects a sparse concept subset in one step and Arthur predicts from this subset, rather than building up a sequential history. Second, \method operates on a fixed, general-purpose concept vocabulary derived from CLIP/LAION captions or extracted unsupervised by NCB, rather than on raw image patches or hand-crafted semantic queries. These differences make the two methods complementary rather than directly substitutable: V-IP produces an interpretable decision chain whose length adapts per input, while \method produces a fixed-size, inspectable concept subset.

\paragraph{CIFAR-100 comparison.} On CIFAR-100, which overlaps with our experiments, V-IP attains approximately $60\%$ classification accuracy after $7$ patch queries on the $8\times 8$ stride-$4$ vocabulary (cf.\ Figure~5b in~\citet{chattopadhyay2023variational}). The CLIP-based instantiation of \method on CIFAR-100 reaches $83.32\%$ completeness with a single-shot concept selection of fixed size~$32$ (\autoref{tab:dcr_comparison}). We stress that this is not a controlled head-to-head experiment: V-IP and \method operate on different feature spaces (pixel patches vs.\ concept similarities) and emit different forms of explanation (sequential queries vs.\ a sparse concept subset).

\paragraph{Scaling considerations.} The size of V-IP's query set scales with the number of candidate features. For pixel-patch queries on $224\times 224$ images with $1{,}000$ classes (\eg, ImageNet-1k), the candidate set grows substantially and individual patches carry less label information, which makes pure pixel-level V-IP challenging in this regime. Semantic-query variants of V-IP avoid this issue but typically rely on a hand-crafted, task-specific query set. \method instead reuses the same general-purpose concept vocabulary across datasets and delegates the difficult selection problem to the Merlin--Morgana--Arthur game, which scales to ImageNet-1k under the same setup used for CIFAR-100.

\section{NCB-based \methodfull Experiments}\label{app:ncb_ncv}

In the following, we provide details on NCB-based \method, experimental evaluations as well as additional evaluations.

\subsection{Pretraining}

Before training \method, we first pretrain the models without the feature selectors. The corresponding results for the pretraining are shown in \autoref{tab:pretraining_clevr_hans}, where we evaluate on 20 random seeds. These pretrained models are then used as initialization for the subsequent \method training. For the pretraining, we use a Set Transformer with two stacked multi-head attention blocks, a hidden dimension of 128 and four attention heads. We use a batch size of 128, 30 epochs and the Adam optimizer with a learning rate of $10^{-3}$ for both datasets, applying early stopping based on validation loss.

\begin{table}[h!]
    \centering
    \caption{Pretraining results on the CLEVR-Hans3 and CLEVR-Hans7 datasets without shortcuts} \vspace{0.5em}
    \begin{tabular}{ c c c c }
    \toprule
    \multicolumn{2}{c}{\textbf{CLEVR-Hans3}} & \multicolumn{2}{c}{\textbf{CLEVR-Hans7}} \\
     Val. Accuracy & Test Accuracy & Val. Accuracy & Test Accuracy \\
    \midrule
    $99.02 \pm 0.31$ & $98.13 \pm 0.37$ & $98.08 \pm 0.24$ & $97.83 \pm 0.25$ \\
    \bottomrule
    \end{tabular}
    \label{tab:pretraining_clevr_hans}
\end{table}

\subsection{\method Training}\label{app:training}

For the experiments presented in our main results in \autoref{tab:scalability}, the experimental details for both datasets are as follows:

\textbf{Model Architecture.}
The verifier is implemented as a pretrained Set Transformer consisting of two stacked multi-head attention blocks with hidden dimension 128, four attention heads, and layer normalization. Merlin and Morgana are implemented as independent neural networks, each parameterized by a Set Transformer with two stacked attention blocks with hidden dimensions 256, four attention heads, and layer normalization. The provers receive the full concept slot matrix as input and output a sparse selection mask with exactly 12 nonzero entries (out of 64 total features), indicating the active blocks provided to the verifier.

\textbf{Training Details.}
All components are jointly trained using the Adam optimizer with a learning rate of $10^{-3}$ and weight decay of $10^{-4}$. Models are trained for 50 epochs and a batch size of 512 is used throughout. For the Merlin and Morgana provers, a hard selection constraint is enforced, limiting the number of selected concepts to a fixed budget of 12 block-encodings per sample. To ensure high soundness, we set $\gamma=0.5$, giving equal weight to both feature selector losses in the total loss computation. We train our models using 20 random seeds.

\textbf{Extended Results.}
Additionally, we evaluated the \method framework with varying mask sizes and an alternative model architecture for the feature selectors. The results are presented in \autoref{tab:ncv_unconf-H3} for the CLEVR-Hans3 dataset and \autoref{tab:ncv_unconf_H7} for the CLEVR-Hans7 dataset, where we evaluate both the validation set and the test set. The alternative architecture implements a MLP with two hidden layers and ReLU activation functions for the feature selectors, while maintaining a pretrained Set Transformer as the classifier across all experiments. Our results reveal that the Set Transformer feature selector consistently outperforms the MLP feature selector on the test set, particularly with smaller mask sizes such as 4 and 6. Furthermore, this configuration maintains high completeness (>96\%) and soundness (>99\%), even with a reduced number of selected features.

\begin{table}[t!]
\centering
\caption{Completeness and soundness on the CLEVR-Hans3 dataset without shortcuts for different mask sizes and feature selector architectures. The highlighted values are used for Table \ref{tab:scalability}. 
} \vspace{0.5em}
\begin{tabular}{clcccc}
\toprule
& & \multicolumn{2}{c}{\textbf{Validation}} & \multicolumn{2}{c}{\textbf{Test}}\\
\textbf{Mask Size} & \textbf{Feature Selector}  & \textbf{Completeness} & \textbf{Soundness} & \textbf{Completeness} & \textbf{Soundness} \\
\midrule
\multirow{2}{*}{4} & Set Transformer & $98.35 \pm 0.31$ & $99.88 \pm 0.28$ & $97.69 \pm 0.63$ & $99.82 \pm 0.23$ \\
& MLP & $96.51 \pm 1.18$ & $99.81 \pm 0.27$ & $95.54 \pm 1.37$ & $99.85 \pm 0.28$ \\
\midrule
\multirow{2}{*}{6} & Set Transformer & $98.71 \pm 0.52$ & $99.87 \pm 0.13$ & $98.11 \pm 0.62$ & $99.88 \pm 0.19$ \\
& MLP & $96.21 \pm 0.89$ & $99.96 \pm 0.03$ & $94.78 \pm 1.24$ & $99.97 \pm 0.06$ \\
\midrule
\multirow{2}{*}{12} & Set Transformer  & $99.20 \pm 0.11$ & $100.00 \pm 0.00$ & $\mathbf{98.92 \pm 0.32}$ & $\mathbf{100.00 \pm 0.00}$ \\
& MLP & $99.28 \pm 0.11$ & $99.98 \pm 0.06$ & $98.89 \pm 0.21$ & $99.99 \pm 0.07$ \\
\bottomrule
\end{tabular}
\label{tab:ncv_unconf-H3}
\end{table}

\begin{table}[t!]
\centering
\caption{Completeness and soundness on the CLEVR-Hans7 dataset without shortcuts for different mask sizes and feature selector architectures. The highlighted values are used for Table \ref{tab:scalability}. 
} \vspace{0.5em}
\begin{tabular}{clcccc}
\toprule
& & \multicolumn{2}{c}{\textbf{Validation}} & \multicolumn{2}{c}{\textbf{Test}}\\
\textbf{Mask Size} & \textbf{Feature Selector}  & \textbf{Completeness} & \textbf{Soundness} & \textbf{Completeness} & \textbf{Soundness} \\
\midrule
\multirow{2}{*}{4} & Set Transformer & $96.69 \pm 1.28$ & $99.93 \pm 0.09$ & $96.71 \pm 1.37$ & $99.91 \pm 0.09$ \\
& MLP & $92.63 \pm 1.24$ & $99.89 \pm 0.12$ & $92.71 \pm 1.31$ & $99.87 \pm 0.13$ \\
\midrule
\multirow{2}{*}{6} & Set Transformer & $97.32 \pm 0.42$ & $99.98 \pm 0.02$ & $97.14 \pm 0.51$ & $99.98 \pm 0.02$ \\
& MLP & $95.43 \pm 1.48$ & $99.88 \pm 0.13$ & $95.12 \pm 1.48$ & $99.86 \pm 0.14$ \\
\midrule
\multirow{2}{*}{12} & Set Transformer & $98.13 \pm 0.11$ & $100.00 \pm 0.00$ & $\mathbf{97.89 \pm 0.31}$ & $\mathbf{100.00 \pm 0.00}$ \\
& MLP & $97.41 \pm 1.07$ & $99.99 \pm 0.03$ & $97.01 \pm 0.93$ & $99.99 \pm 0.04$ \\
\bottomrule
\end{tabular}
\label{tab:ncv_unconf_H7}
\end{table}

\subsection{Explanations} \label{apx:ncv_explanations}

\begin{figure}[t!]
    \centering
    \includegraphics[width=.8\textwidth]{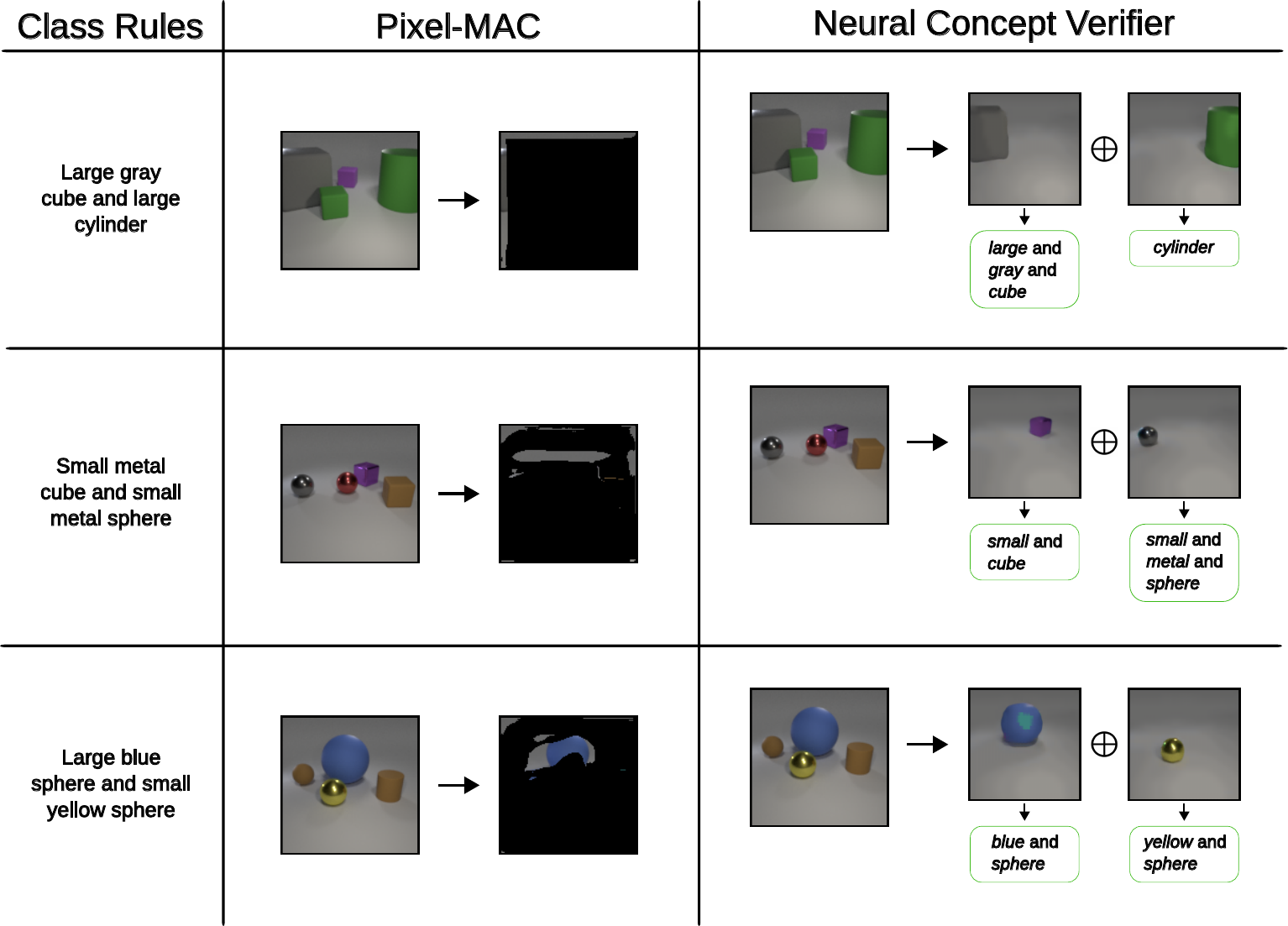}
    \caption{Comparison of explanations from \method \textit{vs.} Pixel-MAC for CLEVR-Hans3 images of all three classes. \textbf{(a)} Merlin–Arthur training on pixel space yields uninformative masks. \textbf{(b)} \method provides clear explanations by highlighting object features corresponding to the class rule. The single-object images are reconstructions from the respective slots selected by Merlin (prover).}
    \label{fig:CLEVRHans3_explanations_allclasses}
\end{figure}

Here, we present supplementary examples of explanations generated by both Pixel-MAC and \method on the CLEVR-Hans3 dataset in \autoref{fig:CLEVRHans3_explanations_allclasses}. These results further substantiate our claim that \method provides significantly more transparent and interpretable explanations compared to the pixel-based PVG baseline.

\section{Experimental details for CLIP-based \method}\label{app:clip_ncv}
In the following section, we present the implementation details of CLIP-based \method training.

\subsection{Pretraining} 

Once more, before starting with the actual \method training, we first pretrain the models without the provers (Merlin and Morgana). The corresponding results for the pretraining are shown in \autoref{tab:pretraining_cifar_imagenet}. As textual concept descriptions $T$, we used the top 10{,}000 most frequent one- and two-word phrases from LAION \citep{schuhmann2021laion} captions, following the setup of \citet{bhalla2024interpreting}. For pretraining the verifier, we use a two-layer multilayer perceptron (MLP) with a hidden dimension of 512 and GELU activations \citep{hendrycks2016gaussian} on CIFAR-100, ImageNet-1k, and COCOLogic. On CIFAR-100 and ImageNet, training uses a batch size of 4096 and a learning rate of $10^{-4}$, with dropout (0.3), weight decay of $10^{-4}$, and early stopping (patience 10). On COCOLogic, we instead train for 100 epochs with a batch size of 512, learning rate of $10^{-4}$, and weight decay of $10^{-2}$, using a learning-rate scheduler (plateau, patience 5, factor $10^{-3}$, minimum learning rate $10^{-6}$) and no early stopping. All pretraining is conducted without provers, and the resulting verifiers are used to initialize the CLIP-based \method training.

\begin{table}[t!]
    \caption{Pretraining accuracy of the verifier (without provers) for CLIP-based \method on CIFAR-100, COCOLogic and ImageNet-1k.} \vspace{0.5em}
    \centering
    \begin{tabular}{lc}
        \toprule
        Dataset & Accuracy (\%) \\
        \midrule
        CIFAR-100 & 85.96 \\
        COCOLogic & 81.39 \\ 
        ImageNet-1k & 77.07 \\
        \bottomrule
    \end{tabular}
    \label{tab:pretraining_cifar_imagenet}
\end{table}

\subsection{\method Training}

For the experiments presented in our main results in \autoref{tab:scalability}, we detail the training setup separately for CIFAR-100 and ImageNet.

\textbf{Model Architecture.}
The verifier (Arthur) is initialized as the pretrained two-layer multilayer perceptron (MLP) described above. Merlin and Morgana are implemented as independent neural networks, each parameterized by a two-layer MLP with hidden dimension 512 and ReLU activations. Both provers receive the full concept activation vector as input and output a sparse selection mask indicating the active concepts that are passed to the verifier.

\textbf{CIFAR-100.}
For CIFAR-100, all components are trained jointly for 100 epochs using the Adam optimizer. We set the verifier learning rate to $10^{-4}$, and use $5 \times 10^{-4}$ for both Merlin and Morgana. A batch size of 256 is used throughout. Weight decay is set to $0.1$, and a hard mask size of 32 concepts is enforced per input. To incentivize sparse masks, an L1 penalty of 0.1 is applied to the provers. A learning rate scheduler (plateau-based) is employed with a patience of 5, minimum learning rate of $10^{-6}$, and decay factor of 0.001. Early stopping is disabled, and all results are averaged over 10 random seeds.

\textbf{ImageNet.}
The ImageNet setup mirrors CIFAR-100 in most aspects. We again train for 100 epochs with a batch size of 256, using the same learning rates for verifier ($10^{-4}$) and provers ($5 \times 10^{-4}$), mask size of 32 features, and L1 penalty (0.1). Weight decay is reduced to $0.005$ to improve generalization. The same learning rate scheduler and seed setup are used as in the CIFAR-100 experiments.

\textbf{COCOLogic.}
Training on COCOLogic follows the CIFAR-100 configuration with minor adjustments: models are trained for 100 epochs with a batch size of 512, verifier learning rate of $10^{-4}$, and prover learning rates of $5 \times 10^{-4}$. We use a weight decay of $0.01$, a mask size of 32, and an L1 penalty of 0.1 on the provers. As with CIFAR-100, early stopping is disabled, and learning rate scheduling and seed averaging remain unchanged.

\begin{table}[t!]
    \centering
    \caption{Ablation results for our method on CIFAR-100 with varying mask sizes. We report mean $\pm$ std over 10 seeds.} \vspace{0.5em}
    \begin{tabular}{lcccc}
        \toprule
        Mask Size & \makecell{Completeness\\Train} & \makecell{Completeness\\Validation} & \makecell{Soundness\\Train} & \makecell{Soundness\\Validation} \\
        \midrule
        4   & $74.97 \std{4.45}$ & $69.32 \std{2.66}$ & $99.85 \std{0.07}$ & $99.85 \std{0.09}$ \\
        8   & $87.42 \std{2.60}$ & $71.20 \std{22.43}$ & $99.95 \std{0.03}$ & $99.95 \std{0.03}$ \\
        16  & $94.08 \std{1.46}$ & $81.82 \std{0.50}$ & $99.96 \std{0.03}$ & $99.97 \std{0.02}$ \\
        64  & $97.65 \std{0.47}$ & $84.01 \std{0.31}$ & $100.00 \std{0.00}$ & $100.00 \std{0.00}$ \\
        \bottomrule
    \end{tabular}
    \label{tab:ablation_masks_cifar}
\end{table}

\begin{table}[t!]
    \centering
    \caption{Ablation results for our method on ImageNet with varying mask sizes. We report mean $\pm$ std over 10 seeds.} \vspace{0.5em}
    \begin{tabular}{lcccc}
        \toprule
        Mask Size & \makecell{Completeness\\Train} & \makecell{Completeness\\Validation} & \makecell{Soundness\\Train} & \makecell{Soundness\\Validation} \\
        \midrule
        4   & $59.30 \std{0.34}$ & $55.96 \std{0.30}$ & $99.81 \std{0.03}$ & $99.83 \std{0.06}$ \\
        8   & $64.84 \std{0.41}$ & $60.98 \std{0.30}$ & $99.94 \std{0.01}$ & $99.94 \std{0.03}$ \\
        16  & $68.85 \std{0.34}$ & $64.60 \std{0.12}$ & $99.96 \std{0.03}$ & $99.96 \std{0.03}$ \\
        64  & $73.35 \std{0.39}$ & $69.03 \std{0.18}$ & $99.97 \std{0.01}$ & $99.97 \std{0.02}$ \\
        \bottomrule
    \end{tabular}
    \label{tab:ablation_masks_imagenet}
\end{table}

\subsection{Prover and Verifier Architectures}
\label{app:arch-choices}

In all experiments, the choice of prover and verifier architectures is guided by the structure of the concept representation. For unordered, slot-based concept encodings (as in NCB on CLEVR-Hans), we use permutation-invariant Set Transformers for Merlin, Morgana, and Arthur. For fixed-size concept vectors (as in CLIP/SpLiCE on CIFAR-100, ImageNet-1k, and COCOLogic-10), we use shallow nonlinear MLPs.

In preliminary experiments, we observed that performance is stable across a range of nonlinear architectures (varying depth, width, and activations), whereas purely linear models consistently underperformed by collapsing to simple correlation tests. Based on this, we recommend using permutation-invariant architectures for slot-based concepts and nonlinear MLPs for vector-based concepts when applying NCV to new datasets.

\subsection{Effect of the weighting parameter $\gamma$}
\label{app:gamma-sweep}

The weighting parameter $\gamma$ controls the trade-off between Merlin's cooperative
objective and Morgana's adversarial objective when training Arthur. Recall that
Arthur's loss is given by
\[
  L_A = (1-\gamma)\,L_M + \gamma\,L_{\widehat{M}},
\]
so that $\gamma = 0$ corresponds to training Arthur only on Merlin's loss, and larger
$\gamma$ increases the relative weight of Morgana's adversarial objective.
To assess the influence of this trade-off on NCV, we sweep $\gamma$ over a range
from $0$ (Merlin-only) up to $0.5$ (substantial weight on Morgana) on CIFAR-100,
ImageNet-1k, and COCOLogic, keeping all other hyperparameters fixed. For each
setting, we train three models with different random seeds and report the mean
completeness and soundness; for COCOLogic, we report balanced metrics due to
class imbalance.

Across all datasets the same qualitative behavior emerges. When $\gamma = 0$,
i.e., the verifier is trained without adversarial pressure from Morgana,
completeness remains high but soundness collapses (e.g., $37.9\%$ on CIFAR-100,
$10.4\%$ on ImageNet-1k, and $51.8\%$ balanced soundness on COCOLogic). As soon
as $\gamma > 0$, soundness rapidly recovers to values close to those reported in
\autoref{tab:scalability}, while completeness remains essentially unchanged
throughout the range of $\gamma$ we consider. These results demonstrate that
incorporating adversarial concept selection is crucial for learning verifiers that
remain reliable under misleading concept subsets, without sacrificing standard
predictive performance in the regimes we study. For a detailed discussion of why
an adversarial prover and its relative strength are essential for the
Merlin--Arthur mutual-information guarantees underlying NCV, see
\autoref{app:theory}.

\begin{figure*}[t]
    \centering
    \begin{minipage}[t]{0.32\textwidth}
        \centering
        \includegraphics[width=\linewidth]{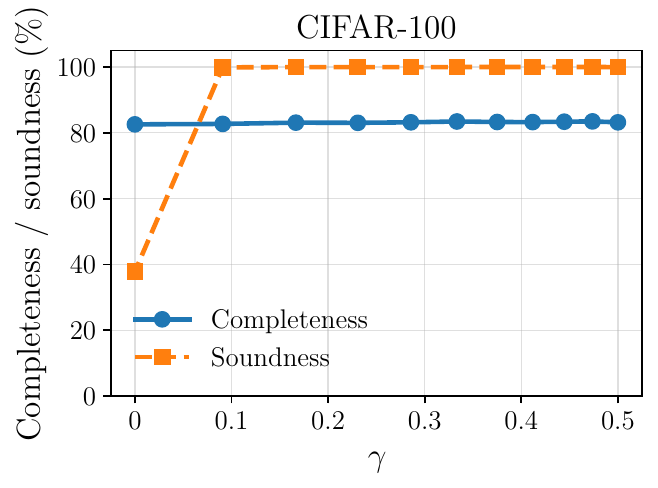}
        \\[0.25em]
        \small (a) CIFAR-100
    \end{minipage}
    \hfill
    \begin{minipage}[t]{0.32\textwidth}
        \centering
        \includegraphics[width=\linewidth]{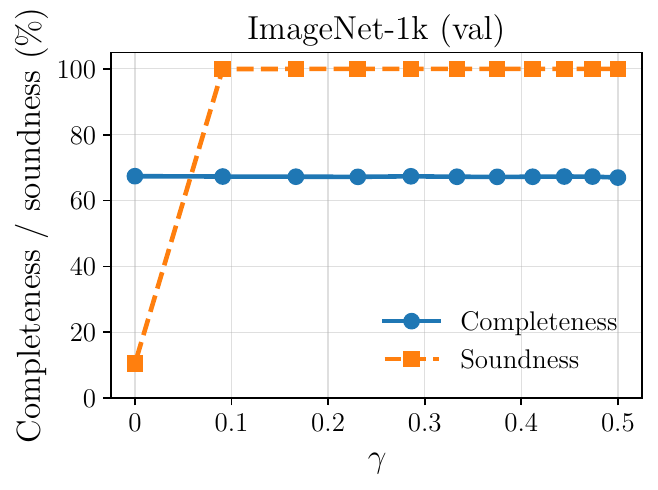}
        \\[0.25em]
        \small (b) ImageNet-1k (validation)
    \end{minipage}
    \hfill
    \begin{minipage}[t]{0.32\textwidth}
        \centering
        \includegraphics[width=\linewidth]{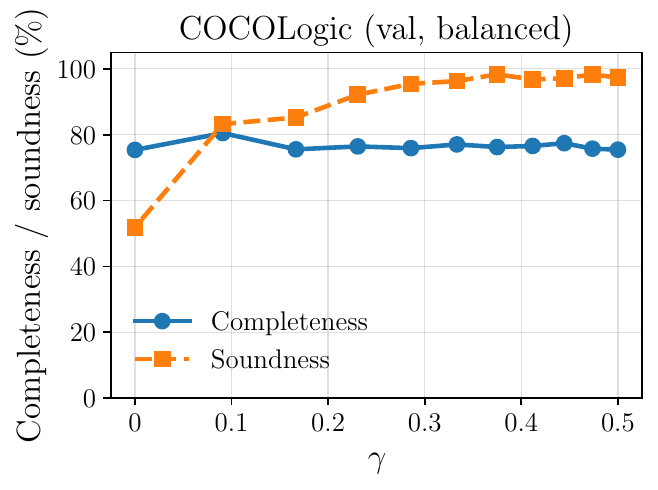}
        \\[0.25em]
        \small (c) COCOLogic (validation, balanced metrics)
    \end{minipage}
    \caption{Effect of the weighting parameter $\gamma$ on completeness and
    soundness for (a) CIFAR-100, (b) ImageNet-1k validation, and (c) COCOLogic
    validation (balanced metrics). All curves show means over 3 random seeds.}
    \label{fig:gamma-sweep}
\end{figure*}

\subsection{Comparison of Selected Concepts: \method{} vs.\ Linear CBM}
\label{app:concept-comparison}

To complement the qualitative comparison in the main text (Q3), we directly compare the concepts selected by \method{} against those of a linear CBM trained on the \emph{same} CLIP/SpLiCE concept vocabulary. \autoref{tab:concept-comparison} reports, for six example ImageNet-1k classes, the five most frequently selected concepts across $32$ test samples per class. For these classes, both methods recover semantically coherent and largely overlapping concept sets, illustrating that \method{}'s sparse, adversarially trained selection can be as meaningful as that of a linear concept predictor. Within a class, each image still receives its own explanation: core concepts (e.g., \emph{fedora} and \emph{rodeo} for \emph{cowboy hat}) appear in roughly $30$ of $32$ samples, while the remaining slots vary per image.

We emphasize that these are illustrative example classes, chosen to compare the two methods on cases where both select clearly meaningful concepts; they are not intended as a claim of uniform behavior across all $1000$ ImageNet-1k classes, where the interpretability of the selected concepts varies. We therefore present this comparison as a qualitative illustration on selected classes rather than a dataset-wide claim about \method{}'s concept selection.

\begin{table}[t]
\centering
\small
\caption{Top-5 most frequently selected concepts across $32$ test samples per class on ImageNet-1k, comparing a linear CBM and \method{} on the shared CLIP/SpLiCE vocabulary derived from LAION captions~\citep{schuhmann2021laion}. Both methods identify semantically similar concepts for each class.}
\label{tab:concept-comparison}
\begin{tabular}{@{}l p{0.40\linewidth} p{0.40\linewidth}@{}}
\toprule
\textbf{Class} & \textbf{Linear CBM (top-5)} & \textbf{\method{} (top-5)} \\
\midrule
cowboy hat & cowboy, fedora, hats, cowgirl, cowboys & fedora, cowgirl, rodeo, southwestern, boots \\
crate & crates, crate, shipment, pallet, wooden & shipment, logistics, crate, wood, cardboard \\
paper towel & wipes, dispenser, scroll, napkin, tissue & tissue, napkin, papers, wipe, toilet \\
candle & candle, candles, soothing, birthday, wei & candle, candles, advent, yen, cover \\
slot machine & slots, slot, jackpot, gained, casino & slots, bet, customizable, odds, bingo \\
web site & homepage, browse, website, websites, newsletter & homepage, websites, screenshot, browse, invite \\
\bottomrule
\end{tabular}
\end{table}

We note that several selected concepts share their surface form with class names (e.g., \emph{crate}, \emph{candle}, \emph{website}). This is a natural consequence of using a general-purpose vocabulary, namely the $10{,}000$ most frequent one- and two-word phrases from LAION captions~\citep{schuhmann2021laion}, from which class names were deliberately not filtered. Crucially, neither \method{} nor the linear CBM receives class labels directly: both operate only on per-image cosine similarities between the CLIP image embedding and all $10{,}000$ concept embeddings, from which Merlin must still select an informative sparse subset. The verifier is therefore not simply handed the answer; producing meaningful similarity scores and selecting an informative subset remains non-trivial even when some concepts share their names with class labels.

\section{Computational Cost and Hardware Setup}
\label{app:computational-cost}

To complement the main results, we report approximate training times for all models and datasets considered in \autoref{sec:experimental_eval}. All runs were executed on a single NVIDIA A100 or A40 GPU; times are reported as rounded wall-clock estimates and are meant to convey relative cost rather than exact benchmarks. For CBM baselines, we \emph{exclude} the per-sample SpLiCE optimization step at inference time, which would add substantial overhead and further increase their deployment cost.

\begin{table}[h!]
    \centering
    \caption{Approximate training time comparison across models and datasets. Times are rounded wall-clock estimates on a single NVIDIA A100 or A40 GPU. For CBM baselines, the per-sample SpLiCE optimization at inference time is omitted here, but would incur significant additional cost. (ImgNet = ImageNet-1k, COCO-10 = COCOLogic-10, CL-H3/CL-H7 = CLEVR-Hans3/7.)}
    \label{tab:runtime_approx}
    \resizebox{\linewidth}{!}{
    \begin{tabular}{lccccc}
        \toprule
        \textbf{Model} 
            & \textbf{CIFAR} 
            & \textbf{ImgNet} 
            & \textbf{COCO-10} 
            & \textbf{CL-H3} 
            & \textbf{CL-H7} \\
        \midrule
        ResNet (baseline) 
            & $\sim$25--30m 
            & $\sim$1.5d 
            & $\sim$50m 
            & $<10$m 
            & $<10$m \\
        Pixel-MAC 
            & $\sim$1.3d 
            & $\sim$3d 
            & $\sim$4h 
            & $<2$h 
            & $<2$h \\
        CBM 
            & $\sim$10--15m 
            & $\sim$4h 
            & $\sim$2m 
            & $\sim$2m 
            & $\sim$2m \\
        \textbf{NCV (ours)} 
            & $\sim$20m\;\;{\footnotesize(11m A + 7m PVG)} 
            & $\sim$5h\;\;{\footnotesize(2.5h A + 2.5h PVG)} 
            & $\sim$5m 
            & $\sim$1.5m 
            & $\sim$5m \\
        \bottomrule
    \end{tabular}
    }
\end{table}

Overall, NCV is $1$--$3$ orders of magnitude cheaper to train than Pixel-MAC across all datasets, while remaining comparable to standard CBMs in runtime. Since concept encodings are precomputed once and reused across models, NCV scales similarly to a conventional classifier without additional architectural overhead, and remains practical even in large-scale settings such as ImageNet-1k.

\section{Use of Large Language Models}
Large language models were used to support this work by assisting with text refinement, implementation of code components (including methods and plot generation), and by providing input during idea development and approach refinement.

%%%%%%%%%%%%%%%%%%%%%%%%%%%%%%%%%%%%%%%%%%%%%%%%%%%%%%%%%%%%%%%%%%%%%%%%%%%%%%%
%%%%%%%%%%%%%%%%%%%%%%%%%%%%%%%%%%%%%%%%%%%%%%%%%%%%%%%%%%%%%%%%%%%%%%%%%%%%%%%

\end{document}